\pdfoutput=1
\documentclass[11pt]{article}
\usepackage[dvipsnames,table]{xcolor}
\usepackage[]{acl}
\usepackage{times}
\usepackage{latexsym}
\usepackage[T1]{fontenc}
\usepackage[utf8]{inputenc}
\usepackage{inconsolata}
\usepackage{pifont}
\usepackage[toc,page]{appendix}
\usepackage{array,url,booktabs,microtype,multirow,hyperref,subcaption,graphicx,mathtools,epstopdf,xspace,amsthm,amsfonts,nicefrac,xcolor,makecell,adjustbox,pifont, float}
\usepackage{purudefs}

\newtheorem{prop}{Proposition}

\newcommand{\ie}{\textit{i}.\textit{e}., }
\newcommand{\eg}{\textit{e}.\textit{g}., }

\newcommand{\alg}{PRIME\xspace}
\newcommand{\algpp}{PRIME++\xspace}

\renewcommand{\do}[1]{\textcolor{cyan}{#1}}

\definecolor{arsenic}{rgb}{0.23, 0.27, 0.29}

\definecolor{silver}{rgb}{0.75, 0.75, 0.75}

\title{Prototypical Extreme Multi-label Classification with a Dynamic Margin Loss}

\author{
  Kunal Dahiya$^{*\dag}$ \\
  IIT Delhi \\
  \small \texttt{kunalsdahiya@gmail.com} \\
  \And
  Diego Ortego$^*$ \\
  NielsenIQ \\
  \small \texttt{diego.ortego@nielseniq.com} \\
  \And
  David Jiménez \\
  NielsenIQ \\
  \small \texttt{david.jimenez@nielseniq.com}
}

\begin{document}
\maketitle
\renewcommand{\thefootnote}{\fnsymbol{footnote}}
\footnotetext[1]{Equal contribution}
\footnotetext[2]{Work done as an intern at NielsenIQ}
\begin{abstract}
Extreme Multi-label Classification (XMC) methods predict relevant labels for a given query in an extremely large label space. 
Recent works in XMC address this problem using deep encoders that project text descriptions to an embedding space suitable for recovering the closest labels.
However, learning deep models can be computationally expensive in large output spaces, resulting in a trade-off between high performing brute-force approaches and efficient solutions. In this paper, we propose \alg, a XMC method that employs a novel prototypical contrastive learning technique to reconcile efficiency and performance surpassing brute-force approaches.
We frame XMC as a data-to-prototype prediction task where label prototypes aggregate information from related queries.
More precisely, we use a shallow transformer encoder that we coin as Label Prototype Network, which enriches label representations by aggregating text-based embeddings, label centroids and learnable free vectors.
We jointly train a deep encoder and the Label Prototype Network using an adaptive triplet loss objective that better adapts to the high granularity and ambiguity of extreme label spaces.
\alg achieves state-of-the-art results in several public benchmarks of different sizes and domains, while keeping the model efficient.
\end{abstract}

\section{Introduction}
\label{sec1:intro}
Extreme Multi-label Classification (XMC) is the task of predicting the most relevant subset of labels in an extremely large label space (potentially millions of labels) for a given query data point. XMC methods find applications in various real-world problems including product recommendation in e-commerce, document tagging, and sponsored search. The label is typically endowed with a short description in such applications. For instance, in a product label space, \textit{"Kerplunk!: Stories"} and \textit{"The Good Samaritan Strikes Again"} are relevant labels for the query \textit{"The Grasshopper Trap"}. In this example, relevancy is defined by products that were seen or bought together, while other applications might define different relevancy relations. The complete example is presented in Table \ref{tab:PRIMEpredsExample}. Note that all the relevant relations are often not available in the ground truth in such a large space, \ie there are missing labels.

\begin{figure}[t]
    \centering
    \includegraphics[width=0.30\paperwidth]{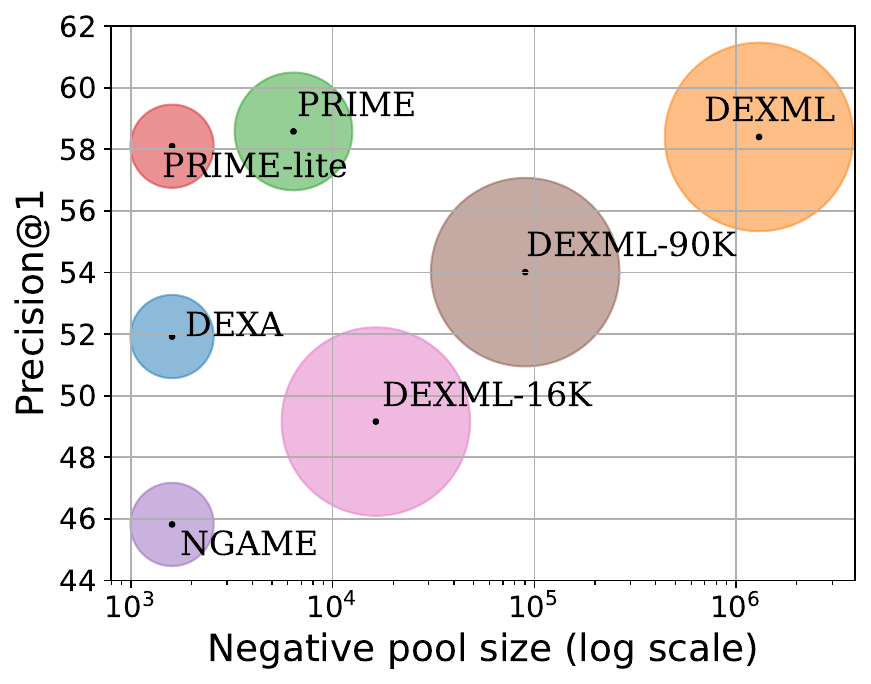}
    \caption{Performance vs efficiency comparison for several encoder-only methods in LF-AmazonTitles-1.3M dataset and our \alg proposal. Blob size represents the models' batch size, y-axis performance, and x-axis number of negatives per query. Note that different versions of DEXML and PRIME vary the negative pool and the batch size, which dominate the method's efficiency.}
    \label{fig:enter-label}
\end{figure}

XMC algorithms typically learn an encoder and extreme classifiers either jointly~\cite{Jain23} or separately by following a modular approach~\cite{Dahiya21, Zhang21}.
This modular approach allows for effortless utilization of label metadata in the form of text~\cite{Dahiya21b}, images~\cite{Mittal22}, or graphs~\cite{Saini21}, thereby enhancing encoder accuracy.
Therefore, learning a robust encoder has become a cornerstone for XMC methods.

Several works in the recent XMC literature focus on learning robust deep encoders using metric learning style of training \cite{Dahiya23, Dahiya23b, Gupta24, Mohan24}. Most of these methods follow a data-to-data metric learning approach, meaning that query and label sentence embeddings are directly obtained from their text descriptions. Despite being possible to get top performance through a data-to-data brute-force approach~\cite{Gupta24} that incorporates all negative labels per query, this strategy is prohibitive in most real-world settings. In particular, this approach incurs a cost of $\bigO{KL}$ to compute query or label embeddings, where $L$ is the number of labels, and $K$ is the complexity of encoding the sentence embeddings (please see section~\ref{app:complexity} in the Appendix for further details). We stress that XMC algorithms should be designed both to achieve top-performance and scale well to large datasets. 

\begin{table}
\begin{centering}
\resizebox{\columnwidth}{!}{
\begin{tabular}{|c|l|}
\hline 
\textbf{Query} & \textbf{Positive labels ($\times 7$)}\tabularnewline
\hline 
 & "How I Got This Way", \tabularnewline
 & "Circles in the Snow: A Bo Tully Mystery",\tabularnewline
"The Grasshopper & "With Recipes and Commentaries", \tabularnewline
Trap" & "Real Ponies Don't Go Oink!",\tabularnewline
 & "The Good Samaritan Strikes Again",\tabularnewline
 & "Kerplunk!: Stories.", "The Bear in the Attic"\tabularnewline
\hline
\hline
\multicolumn{2}{|c|}{\textbf{DEXA predictions ($\times 5$)}}\tabularnewline
\hline 
\multicolumn{2}{|c|}{\textcolor{red}{"The Ant and the Grasshopper: An Aesop's Fable"}, }\tabularnewline
\multicolumn{2}{|c|}{\textcolor{red}{"Ant and Grasshopper"}, \textcolor{red}{"Grasshopper" \textit{(ID1)}},}\tabularnewline
\multicolumn{2}{|c|}{\textcolor{red}{"Grasshopper" \textit{(ID2)}}, \textcolor{red}{"Grasshopper" \textit{(ID3)}}.}\tabularnewline
\hline 
\multicolumn{2}{|c|}{\textbf{PRIME predictions ($\times 5$)}}\tabularnewline
\hline 
\multicolumn{2}{|c|}{\textcolor{ForestGreen}{"Kerplunk!: Stories"}, \textcolor{ForestGreen}{"The Good Samaritan Strikes Again"}}\tabularnewline
\multicolumn{2}{|c|}{\textcolor{red}{"The Tamarack Murders: A Bo Tully Mystery"},}\tabularnewline
\multicolumn{2}{|c|}{\textcolor{ForestGreen}{"How I Got This Way"}, \textcolor{ForestGreen}{"The Bear in the Attic"}.}\tabularnewline
\hline 
\end{tabular}
}
\par\end{centering}
\caption{Qualitative example from LF-AmazonTitles-1.3M dataset. PRIME recovers semantically related products, while DEXA predicts irrelevant ones containing some query words. Red and green indicate incorrect and correct predictions, respectively.} \label{tab:PRIMEpredsExample}
\end{table}

In this paper, we introduce a novel method called PRIME, which efficiently learns a robust encoder by leveraging data-to-prototype relations.
We are inspired by works demonstrating that embedding representations from groups of data points can narrow down metric learning complexity~\cite{Snell17, Dopierre21, Kim20}. In particular, we compute label embeddings using a novel Label Prototype Network that learns to aggregate text-based embeddings, label centroids and learnable free vectors. We refer to the proposed aggregated representation as a label prototype as multiple queries are used for its estimation.

Additionally, for better adaptation to the variable hardness of positive and negative labels, we propose to incorporate a dynamic margin in our training loss objectives. We empirically show that this margin is a simple yet effective mechanism to boost XMC performance and theoretically demonstrate its effect (see Appendix~\ref{sec:appendix}). More precisely, the proposed dynamic margin enables positive and negative labels to be projected closer in the embedding space when they are semantically related, while also considering uncertain cases with ambiguous samples or missing labels (a well known limitation in extreme label spaces).
As a result, \alg can recover semantically related products more accurately than related methods (we refer the reader to Table~\ref{tab:PRIMEpredsExample} for a visual example). Furthermore, our experiments demonstrate that \alg achieves state-of-the-art results within a single GPU budget and outperforms brute-force approaches defined in DEXML~\cite{Gupta24} (encoder-only) and Renée~\cite{Jain23} (extreme classifier-based). Note that solutions like DEXML use up to 16$\times$A100 GPUs in some configurations.

Our main contributions are as follows:
\begin{itemize}
    \item We propose a high performing yet efficient XMC encoder-based method based on label prototypes that aggregate information from multiple queries. 
    \item Our multi-objective optimization incorporates a novel adaptive triplet loss formulation that accounts for high granularity and uncertainty, inherent to extremely large label spaces.
    \item \alg yields large performance improvements when compared with methods requiring comparable resources (see Figure~\ref{fig:enter-label}). \alg notably outperforms brute-force approaches in all experiments and for most metrics, while maintaining efficiency. 
\end{itemize}

\section{Related work}
\label{sec2:prev_work}

\textbf{Extreme Multi-label Classification (XMC).} Deep XMC methods~\cite{You18,Kharbanda22,Chien23,Dahiya23b,Gupta24} have demonstrated that learning task-specific embeddings can yield significantly more accurate results as compared to traditional methods (see Appendix~\ref{app:related} for details). It is well established that jointly learning the encoder and the extreme classifiers can be expensive with a large number of labels~\cite{Dahiya21}. In particular, the brute-force approach employed by Renée~\cite{Jain23} can lead to accurate results, however, it is expensive to scale beyond a million labels even with multiple GPUs.
Negative sampling approaches, i.e. ANCE~\cite{Xiong20} and NGAME~\cite{Dahiya23} narrow down the training complexity by selecting a small subset of hard negative labels. Moreover, several methods use a modular strategy~\cite{Dahiya21,Zhang21} that decouples the training of encoders and classifiers to improve scalability. ECLARE~\cite{Mittal21b}, PINA~\cite{Chien23}, and XR-Transformers~\cite{Zhang21b} train the encoder by learning meta-classifiers, \ie classifiers are learned for groups of labels. Conversely, NGAME~\cite{Dahiya23b} proposed to learn a shared deep encoder in a Siamese fashion, where labels are represented via label embeddings. Building on NGAME, DEXA~\cite{Dahiya23b} introduces auxiliary parameters to bridge the semantic gap, \ie the lack of information to accurately represent labels when their descriptions are short. On the other hand, DEXML \cite{Gupta24} demonstrates that Siamese deep encoders alone can theoretically yield state-of-the-art results using a brute-force approach with large batch sizes and complete label sets as negatives.

\textbf{Deep metric learning (DML).} Contrastive learning using the vanilla contrastive loss \cite{LeCun05}, triplet loss \cite{Schroff15} or InfoNCE losses \cite{Vinyals18} is the core of DML across NLP and computer vision. In NLP, they are widely used to learn robust sentence representations \cite{Gao21, Zhang21b} or, more generally, multi-purpose embeddings with encoders \cite{Wang22} or decoder-only architectures \cite{Muennighoff24, Wang24}. As for XMC, DML has successfully been applied using deep encoders~\cite{Dahiya23, Dahiya23b, Mohan24, Gupta24}. However, this approaches are based on data-to-data comparisons, which in XMC becomes specially complex, as demonstrated by the great benefits of increasing batch-wise comparisons \cite{Gupta24}. However, the DML community has proposed alternatives to narrow down such complexity exploiting data-to-prototype \cite{Zeng22} or data-to-proxy \cite{Kim20} relations. Prototypes and proxies are both representative embeddings for groups of semantically similar instances \cite{Kim20, Zeng22}. The difference is that proxies are usually learned as network parameters \cite{Kim20, Ren24}, while prototypes are computed averaging embeddings of related instances \cite{Snell17, Dopierre21, Song22, Zeng22}. Therefore, a few set of proxies or prototypes can capture the global structure of an embedding space and replace data-to-data comparisons for reduced training complexity \cite{Kim20}.

\textbf{Adaptive triplet loss.} The triplet loss \cite{Schroff15} is a widely used alternative for metric learning across many fields \cite{Schroff15, Liu21, Nguyen22}. This loss function ensures that anchor-positive label relations have higher similarity than anchor-negative label ones, while including a margin to force a sufficiently high similarity gap. Despite methods in XMC using a fixed margin that lacks flexibility to address ambiguous situations \cite{Dahiya23, Dahiya23b, Mohan24}, some works in the wider metric learning community have proposed dynamic margins to tackle this limitation. For example, \cite{Liu21} propose a margin function for measuring more detailed similarities between taxonomy paths within the context of self-supervised taxonomy expansion. In forensic medical image matching, authors in \cite{Nguyen22} propose an auto-margin strategy based on exploiting similarity statistics over time. Similarly, \cite{So23} work improves the triplet loss by adapting the margin based on the interpolation strength of mixed samples.

\section{Method}
\label{sec3:method}
\begin{figure}[t]
\centering{}\includegraphics[width=1.0\columnwidth]{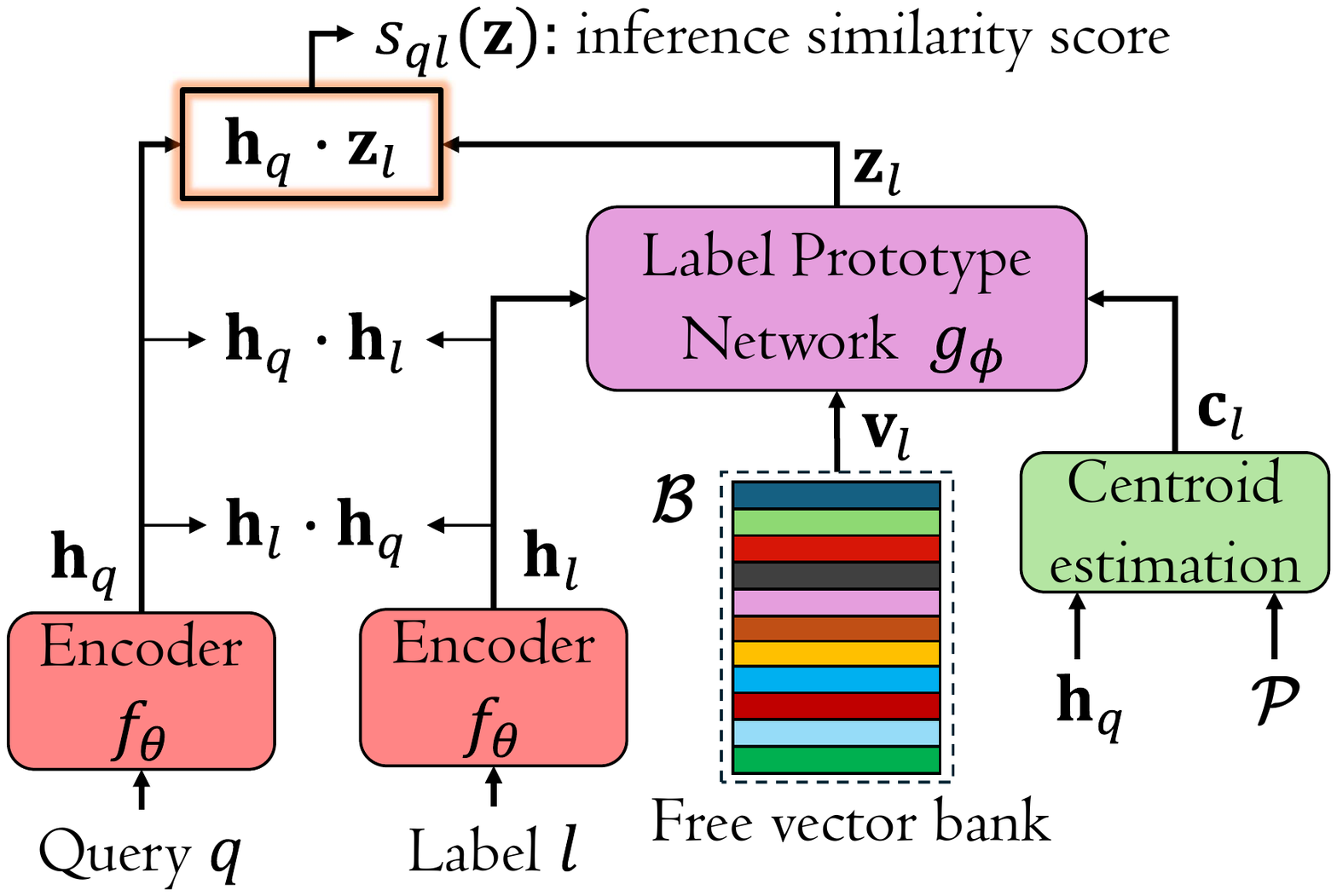}
\caption{Overview of PRIME (PRototypIcal extreME multi-label classification), which exploits query-to-prototype, query-to-label and label-to-query relations.}
\label{fig:method}

\end{figure}

We propose \alg, a novel method for XMC based on label prototypes rather than the usual text-based label embeddings from related works~\cite{Dahiya23, Gupta24, Mohan24}. \alg computes data-to-prototype relations and uses a dynamic margin in its triplet loss objectives, achieving state-of-the-art results on multiple benchmarks. The overview of \alg's architecture is presented in Figure~\ref{fig:method}.

\paragraph{Setup:} Consider~$\cD = \{q_i, \cP_i\}_{i=1}^{Q}$ be a training set with $Q$ queries, where each query $q_i$ has a positive label set $\cP_i$. We define the space of L labels $\cY=\{l_r\}_{r=1}^{L}$ for any $q_i$ to be a super-set of the positive $\cP_i = \{p_j\}_{j=1}^{P_i}$ and the negative sets $\cN_i = \{n_k\}_{k=1}^{N_i}$. $P_i$ and $N_i$ are the number of positive and negative labels associated to $q_i$, respectively. In practice, we differentiate two types of approaches depending on the definition of $\cN_i$: (i) brute-force, which use the complete set of negatives for every query~\cite{Gupta24}; (ii) negative mining, which sample a small subset of hard negative labels during training to strike a balance between accuracy and efficiency~\cite{Xiong20, Dahiya23}.

We aim to learn a function $f_{\theta}: \cQ \rightarrow \cY$ to accurately predict the positive labels for a query $q_i$, where $\theta$ denotes the parameters of a neural network model.
This function encodes the given textual representation into a $d$-dimensional sentence embedding and is used to encode every query $q_i$ and label $l_r$ into $\vh_q^i$ and $\vh_l^r$. Note that $\vh_{l}$ can be defined as $\vh_{p}$ and $\vh_{n}$ to distinguish between positive and negative text-based label embeddings, respectively. 

Existing Siamese training for XMC~\cite{Dahiya23, Mohan24} directly imposes a data-to-data triplet loss objective as follows:
\begin{equation}
\mathcal{L}_T=\sum_{i=1}^{B}\underset{\substack{j\in\mathcal{P}_i\\
k\in\mathcal{N}_i
}
}{\sum}\max(\vh_{q}^{i}\cdot \vh_{n}^{k}-\vh_{q}^{i}\cdot \vh_{p}^{j}+m, 0),\label{Eq:TripletLoss}
\end{equation}
where $B$ is the number of batch queries, $m$ is a fixed margin and $\vh$ refers to L2-normalized embeddings. We simplify the notation in Eq.~\ref{Eq:TripletLoss} and refer to $\cL_T(\vh_q, \vh_l)$ as $\cL_T$. Also, Eq.~\ref{Eq:TripletLoss} introduces abuse of notation when using the set of negative labels $\cN_i$ as the subset of negatives used for each query.

For simplicity, in the remaining of the paper we remove the superscript for query and label indexes and re-define $\cL_T$ using similarity scores as:
\begin{eqnarray}
    \nonumber
    {\cL}_T & = & \max{\left(s_{qn} - s_{qp} + m, 0 \right)} \\
    & = & 
    \begin{cases}
        \Delta s_{q}^{n-p} + m, & \textrm{if~~} \Delta s_{q}^{p-n} \le m\tabularnewline
        0, & \textrm{otherwise}
    \end{cases},
    \label{Eq:triplet_ang}
\end{eqnarray}
where $s_{qn} = \vh_{q}\cdot \vh_{n}$ and $s_{qp} = \vh_{q}\cdot \vh_{p}$ denote the query-negative and the query-positive cosine similarities, respectively. We denote $\Delta s_{q}^{n - p} = s_{qn} - s_{qp}$ and $\Delta s_{q}^{p-n} = s_{qp} - s_{qn}$ as the differences of the cosine similarities between the query-positive and query-negative tuples, respectively. 

\subsection{Dynamic margin}
The traditional triplet loss formulation presented in Eq.~\ref{Eq:triplet_ang} imposes a fixed margin $m$ during training such that all negatives are pushed apart from the query, while positives are pulled close given the condition $\Delta s_{q}^{p-n}\le m$.
Despite being possible to define margins that work well in practice, not all triplets are equally hard, thus suggesting that fixed values might be suboptimal. To overcome this limitation, we propose a dynamic margin that provides adaptation to each triplet hardness. 

\begin{prop}
\label{prop1}
Consider the non-differentiable piece-wise linear function defining the adaptive margin to be ~$m\left(s_{ap}, s_{an}\right) = | s_{ap} - s_{an}|$. Adding $m\left(s_{ap}, s_{an}\right)$ into Eq.~\ref{eq:app_triplet_ang} expands the support of the function by relaxing the margin constraint of the original triplet formulation, inducing a modulation effect that allows semantically similar representations to be projected closer in the embedding space.
\end{prop}

Intuitively, this modulation of the margin enables a more informative embedding space by accounting for degrees of similarities between positives and negatives, \ie the higher the semantic similarity, the closer they will project. Conversely, for high differences where negatives exhibit a much higher similarity than positives, we impose a large penalty.

\begin{prop}
\label{prop2}
Clipping the values of the dynamic margin partitions the triplet loss landscape allowing positives and negatives in uncertain settings to be projected nearby in the embedding space.
\end{prop}

Consider $\textrm{clip}(x)$ be the clipping function,
{
\begin{equation}
    \nonumber
    \textrm{clip}(x) = 
    \begin{cases}
        x, & \textrm{if}\quad \gamma_{\textrm{min}} \le x \le \gamma_{\textrm{max}} \tabularnewline
        \gamma_{\textrm{min}}, & \textrm{if}\quad x < \gamma_{\textrm{min}} \tabularnewline
        \gamma_{\textrm{max}}, & \textrm{if}\quad x > \gamma_{\textrm{max}}
    \end{cases},
    \label{eq:triplet_clip}
\end{equation}
}and consider the conditions for query-to-label similarities  $\cC_p:s_{qp}>s_{qn}$ and $\cC_n:s_{qn}>s_{qp}$ be the inequality conditions of the dynamic margin defined in Proposition~\ref{prop1}. Then, the triplet loss with clipped dynamic margin $\cL_T^{cd}$ becomes:

{\small
\begin{align}
    \cL_T^{cd} &=
    \begin{cases}
        0, & \textrm{if~~} \cC_p, ~~\Delta_q^{p-n} \ge \gamma_{\textrm{min}} \tabularnewline
        \Delta_q^{p-n} + \gamma_{min}, & \textrm{if~~} \cC_p,~~ \Delta_q^{p-n} < \gamma_{\textrm{min}} \tabularnewline
        \Delta_q^{n-p} + \textrm{clip}\left(\Delta s_{q}^{n-p}\right), & \textrm{if~~} \cC_n
    \end{cases}.
    \label{eq:triplet_clip_dyn}
\end{align}
}

In Eq.~\ref{eq:triplet_clip_dyn} we show that the addition of the dynamic margin removes the fixed margin constraint of Eq.~\ref{Eq:TripletLoss} enabling the model to learn from any observation that satisfies condition $\cC_n$, where negatives are closer to the query than positives. Besides, the proposed formulation introduces a new learning region where the gradient direction is inverted: $ \cC_p,~~ \Delta_q^{p-n} < \gamma_{\textrm{min}}$. We argue that this region, where positives and negatives are nearly equally distant to the query, covers cases with high uncertainty such as fine-grain differences or missing labels, which represent well-known challenges in extreme multi-label setups.
For these triplets, negatives and positives could be, respectively, missing labels and false positives. Therefore, by reverting the learning process in this small margin, the network does not take risks and separates positives, while pulling negatives close to the query. Note that if the negative had higher similarity than the positive, this behaviour would not continue. We refer the reader to Appendix~\ref{sec:appendix} for a detailed analysis with proofs of the propositions.

\subsection{Label prototypes}

We propose a prototypical contrastive learning method aiming at narrowing down data-to-data complexity during training, where we build label prototypes as enhanced label embeddings. We do so by using a Label Prototype Network $g_{\phi}$ that aggregates three sources of information: label-text embeddings $\vh_{l}$, label centroids $\vc_{l}$ and learnable free vectors $\vv_{l}$ (see Figure~\ref{fig:method} for reference).
The resulting aggregation produces the label prototype, $$\vz_{l}=g_{\phi}\left(\vh_{l},\vc_{l},\vv_{l}\right).$$
In particular, we use a transformer encoder block layer~\cite{Vaswani17} for~$g_{\phi}$~to make the three sources interact via self-attention and build the label prototype via mean pooling of the resulting contextualized embeddings.
These label prototypes reduce the complexity of data-to-data relations, as they are estimated from centroids and free vectors, which contain an aggregated information from queries and labels not present when solely using text-based label embeddings.

We compute label centroids using a momentum-based approach that updates its values using the batch queries containing the label at hand by:
\begin{equation}
\vc_{p}\,\leftarrow\,\alpha\,\vc_{p}+\left(1-\alpha\right)\vh_{q}^{i},\label{Eq:EMA}
\end{equation}
where $\alpha$ is the momentum coefficient and $\vh_{q}^{i}$ is the i-th query embedding such that $p\in\mathcal{P}_{i}$. We smoothly update the centroids using a high $\alpha$ value. 

Learnable free vectors are normally used in the literature to model side-information that can be complementary to text descriptions, e.g. for modeling attributes in dense retrieval~\cite{Kong22, Sun24} and when modeling groups of labels~\cite{Dahiya23b} in XMC. We adopt the latter strategy and define a bank of free vectors $\mathcal{B}$ where every free vector $\vv_{l}$ in the bank is shared across multiple semantically similar labels, \ie clusters of labels computed before training as done in \cite{Dahiya23b}. These free vectors are learned during training as gradients flow through query-to-prototype relations, thus providing an additional degree of freedom that does not depend directly on the text of queries or labels. Note that the authors in \cite{Dahiya23b} directly add $\vv_{l}$ to $\vh_{l}$, limiting the contribution of free vectors to a linear combination. Conversely, we integrate free vectors as one of the building components used by our Label Prototype Network, thus learning how to incorporate free vectors into the label prototypes for better performance.

\subsection{Encoder and prototype network training}

We jointly train the encoder and the Prototype Label Network using a multi-objective optimization.
We use the proposed dynamic triplet loss function $\mathcal{L}_T^{cd}(\vh_{q},\vz_{l})$ to optimize query-to-prototype relations, which is how we compute our similarity scores at inference time as shown in Figure~\ref{fig:method}. This objective does not directly focus on the optimization of text-based embeddings $\vh$, which is important for learning a strong backbone. Therefore, we propose to include two additional objectives, $\mathcal{L}_T^{cd}(\vh_{q},\vh_{l})$ and $\mathcal{L}_T^{cd}(\vh_{l},\vh_{q})$, that promote query-to-label and label-to-query relations in the text-based embedding space, respectively. This type of backbone optimization has proved to be effective for dense retrieval \cite{Li21, Mesquita22} as it increases the data-to-data relations exploited during training and learns a better embedding space.
Our final loss is then defined as:
\begin{equation}
\resizebox{.99\columnwidth}{!}{
$\mathcal{L}=\mathcal{L}_T^{cd}(\vh_{q},\vz_{l}) + \mathcal{L}_T^{cd}(\vh_{q},\vh_{l}) + \mathcal{L}_T^{cd}(\vh_{l},\vh_{q}) + \lambda\mathcal{R}$},
\label{Eq:FinalLoss}
\end{equation}
where $\lambda$ is the contribution of the  regularization term $\mathcal{R}$. 
We bring this regularization from authors in~\cite{Mohan24}, which regularize similarities from a cross-attention module that incorporates meta-data information into query embeddings to be better than similarities produced without the meta-data. We apply the same concept in our approach to enforce query-to-prototype similarities to be better than query-to-label ones, as they are enriched with centroids and free vectors.
We define $\mathcal{R}$ as the average of $\mathcal{R}_{p}$ and $\mathcal{R}_{n}$:
\begin{align}
\mathcal{R}_{p}&=\frac{1}{P}\underset{p\in\mathcal{P}_{i}}{\sum}s_{qp}-b_{qp}+m^{\prime},\\
\mathcal{R}_{n}&=\frac{1}{N}\underset{n\in\mathcal{N}_{i}}{\sum}b_{qn}-s_{qn}+m^{\prime},
\end{align}
where $\mathcal{R}_{p}$ and $\mathcal{R}_{n}$ are the positive and negative similarity differences, $m^{\prime}$ is a fixed margin and $b_{ql}=h_{q}\cdot z_{l}$ is the query-to-prototype similarity for label $l$, i.e. a positive $p$ or negative $n$ label.
Note that $\cP_i$ and $\cN_i$ are the number of positive and negative comparisons in the batch, respectively.

\section{Experiments}
\label{sec4:experiments}

\subsection{Datasets and metrics}
We benchmark our method on 6 publicly available XMC datasets~\cite{XMLRepo} (details in Table~\ref{tab:DatasetDetails} of Appendix~\ref{app:datasets}) covering: product recommendation from LF-AmazonTitles-131K, LF-Amazon-131K and LF-AmazonTitles-1.3M; the prediction of relevant Wikipedia categories in LF-WikiTitles-500K and LF-Wikipedia-500K; and predicting similar Wikipedia articles in LF-WikiSeeAlso-320K. Note that these datasets include both short and long-text settings, demonstrating the applicability of \alg in diverse contexts.

Our main focus is to compare with encoder-only approaches:
DPR~\cite{Karpukhin20}, ANCE~\cite{Xiong20}, SiameseXML~\cite{Dahiya21b}, GraphFormer~\cite{Yang21}, NGAME~\cite{Dahiya23}, DEXA~\cite{Dahiya23b} and DEXML~\cite{Gupta24}. However, we also benchmark \alg with classifier methods: XR-Transformer \cite{Zhang21}, LightXML~\cite{Jiang21}, ELIAS~\cite{Gupta22}, CascadeXML~\cite{Kharbanda22}, PINA~\cite{Chien23} and Renée~\cite{Jain23}.

We mainly adopt precision and propensity-scored precision (P@k and PSP@k) metrics defined at~\cite{XMLRepo} for evaluation.

\subsection{Implementation details}
\textbf{Hard negative and positive sampling.} Triplet selection involving hard negatives and positives can have a significant impact on performance. We balance the visualization of positive labels by sampling with a distribution over propensity scores of labels (see Appendix~\ref{app:sampling} for more details). Moreover, we sample two positives per query, unless otherwise stated. Please note that the memory and compute requirements increase with number of sampled positives (see Appendix~\ref{app:complexity} for more details on complexity).
On the other hand, we adopt the efficient NGAME~\cite{Dahiya23} negative sampling strategy, which constructs batches in such a way that inexpensive in-batch sampling offers semi-hard negatives.

\textbf{Training and inference.}
We adopt common practices and configurations from previous works. To keep the comparison fair, we adopt the 66M parameter DistilBERT~\cite{Reimers19} encoder used in other papers and report results for a single run. We compute sentence embeddings by mean pooling token representations.
We use maximum inner product search between queries and label prototype embeddings at inference time. We refer the reader to Appendix \ref{app:hyper_parameters} and \ref{app:inference} for further details.
Finally, we set all PRIME configurations to fit in a single 80 GB A100 GPU.

\subsection{Results}\label{subsec:ResultsComparison}
In tables \ref{tab:results_repo_amazon} and \ref{tab:results_repo_wiki} we present a comparative evaluation of \alg against most relevant encoder-based XMC algorithms on Amazon and Wikipedia datasets, respectively. Our results demonstrate that \alg outperforms recent works (ANCE, DEXA and DEXML) in all experiments and for most metrics. In particular, \alg achieves state-of-the-art performance in the largest and most challenging dataset LF-AmazonTitles-1.3M, outperforming the brute-force approach DEXML using $16\times$ more computational resources, and surpassing by more than 6 points in P@1 the rest of the methods. We extend our comparison with brute force approaches in Subsection~\ref{subsec:PRIMEvsDEXML} (see Figure~\ref{fig:enter-label} for a visual comparison).

\alg's primary objective revolves around efficiency in XMC models and thus our focus on classifier-free encoder-based approaches.
Nevertheless, learning classifiers on top of \alg's embeddings (we follow NGAME's~\cite{Dahiya23} One-vs-All approach) might provide moderate gains over the encoder-only model. For completeness, we present results using extreme classifiers in Table~\ref{tab:results_classifiers} on the largest Amazon and Wikipedia datasets, demonstrating the superior performance of our \alg proposal. Note that Renée can be seen as a counterpart of DEXML as it learns the extreme classifiers in a brute-force fashion, \ie without negative sampling.

The results on all datasets are included in Appendix~\ref{app:app_ova}.

\begin{table}[ht]
      \centering
      \resizebox{0.9\linewidth}{!}{
        \begin{tabular}{@{}l|cccc@{}}
        \toprule
        \textbf{Method} & \textbf{P@1} & \textbf{P@3} & \textbf{P@5} & \textbf{PSP@5}  \\ \midrule

        \multicolumn{5}{c}{\textbf{LF-AmazonTitles-1.3M}}\\ \midrule
        \textbf{DPR} & 44.64 & 39.05 & 34.83 & 36.72 \\
        \textbf{ANCE} & 46.44 & 41.48 & 37.59 & 37.25 \\
        \textbf{SiameseXML} & 43.80 & 38.60 & 34.94 & 28.48 \\
        \textbf{NGAME} & 45.82 & 39.94 & 35.48 & 36.80 \\
        \textbf{DEXA} & 51.92 & 44.01 & 38.86 & \underline{37.31} \\
        \textbf{DEXML}~\ding{70} & \underline{58.40} & - & \textbf{45.46} & 36.58 \\
        \midrule
        \textbf{\alg} & \textbf{58.58} & \textbf{50.83} & \underline{45.44} & \textbf{39.07} \\
        \midrule
        \multicolumn{5}{c}{\textbf{LF-AmazonTitles-131K}}\\ \midrule
        \textbf{GraphFormers} & 20.84 & 13.57 & 10.06 & 24.93 \\
        \textbf{DPR} & 41.85 & 28.71 & 20.88 & 49.45 \\
        \textbf{ANCE} & 42.67 & 29.05 & 20.98 & 49.03 \\
        \textbf{SiameseXML} & 41.42 & 27.92 & \underline{21.21} & 46.19 \\
        \textbf{NGAME} & 42.61 & 28.86 & 20.69 & 48.71 \\
        \textbf{DEXA} & \underline{44.76} & \underline{29.72} & 21.18 & \underline{49.50} \\
        \textbf{DEXML}~\ding{70} & 42.52 & - & 20.64 & 47.40 \\
        \midrule
        \textbf{\alg} & \textbf{44.87} & \textbf{30.06} & \textbf{21.53} & \textbf{49.73} \\
        \midrule
        \multicolumn{5}{c}{\textbf{LF-Amazon-131K}}\\ \midrule
        \textbf{DPR} & 43.30 & 29.74 & 21.90 & \underline{51.52} \\
        \textbf{ANCE} & 44.87 & 30.31 & 21.89 & 50.12 \\
        \textbf{NGAME} & 45.35 & 29.89 & 21.35 & 49.32 \\
        \textbf{DEXA} & \underline{46.64} & \underline{30.93} & \underline{22.06} & 50.38 \\
        \textbf{DEXML}~\ding{70} & - & - & - & - \\
        \midrule
        \textbf{\alg} & \textbf{48.09} & \textbf{32.39} & \textbf{23.34} & \textbf{53.43} \\
        \bottomrule
    \end{tabular}
    }
    \caption{Classifier-free evaluation on product recommendation.\ding{70} denotes brute-force algorithm. Bold and underlined numbers denote best and second best results, respectively. `-' indicates results not available.
    }
    \label{tab:results_repo_amazon}
\end{table}

\begin{table} [t]
      \centering
      \resizebox{0.9\linewidth}{!}{
        \begin{tabular}{@{}l|cccc@{}}
        \toprule
        \textbf{Method} & \textbf{P@1} & \textbf{P@3} & \textbf{P@5} & \textbf{PSP@5}  \\ \midrule

        \multicolumn{5}{c}{\textbf{LF-Wikipedia-500K}}\\ \midrule
        \textbf{GraphFormers} & 31.10 & - & 14.00 & 21.83 \\
        \textbf{DPR} & 65.23 & 45.85 & 35.23 & 49.90 \\
        \textbf{ANCE} & 63.33 & 43.35 & 33.12 & 39.71 \\
        \textbf{SiameseXML} & 50.33 & 32.81 & 24.86 & 32.51 \\
        \textbf{NGAME} & 77.92 & 54.87 & 40.95 & 57.33 \\
        \textbf{DEXA} & 79.99 & 57.08 & 42.52 & 57.68 \\
        \textbf{DEXML}~\ding{70} & \textbf{85.78} & - & 50.53 & \textbf{58.97} \\
        \midrule
        \textbf{\alg} & \underline{85.37} & \textbf{65.56} & \textbf{50.92} & \underline{57.74} \\
        \midrule
        \multicolumn{5}{c}{\textbf{LF-WikiTitles-500K}}\\ \midrule
        \textbf{GraphFormers} & 24.53 & - & 11.33 & 19.53 \\
        \textbf{ANCE} & 29.68 & - & 12.51 & 21.18 \\
        \textbf{NGAME} & 29.68 & 18.06 & 12.51 & 21.18 \\
        \textbf{DEXA} & \underline{34.76} & \underline{20.88} & \underline{14.39} & \underline{23.83} \\
        \textbf{DEXML}~\ding{70} & - & - & - & - \\
        \midrule
        \textbf{\alg} & \textbf{46.33} & \textbf{25.98} & \textbf{18.09} & \textbf{24.03} \\
        \midrule
        \multicolumn{5}{c}{\textbf{LF-WikiSeeAlso-320K}}\\ \midrule
        \textbf{DPR} & 41.66 & 27.16 & 20.66 & 36.25 \\
        \textbf{ANCE} & 44.35 & 29.15 & 21.99 & 37.15 \\
        \textbf{SiameseXML} & 40.70 & 27.16 & 20.74 & 35.67 \\
        \textbf{NGAME} & 43.58 & 28.01 & 20.86 & 36.03 \\
        \textbf{DEXA} & \underline{46.57} & \underline{29.92} & \underline{22.26} & \underline{38.27} \\
        \textbf{DEXML}~\ding{70} & - & - & - & - \\
        \midrule
        \textbf{\alg} & \textbf{48.00} & \textbf{31.41} & \textbf{23.53} & \textbf{40.20} \\        
        \bottomrule
    \end{tabular}
    }
    \caption{Classifier-free evaluation on  Wikipedia category and similar page prediction.\ding{70} denotes brute-force algorithm and bold and underlined best and second best. `-' indicates that results are not available.
    }
    \label{tab:results_repo_wiki}
\end{table}

\begin{table}[t]
\resizebox{1.0\columnwidth}{!}{
    \begin{tabular}{@{}l|cccc}
        \toprule
        \textbf{Method}  & \textbf{P@1} & \textbf{P@5} & \textbf{PSP@1} & \textbf{PSP@5} \\ \midrule
         \multicolumn{5}{c}{\textbf{LF-AmazonTitles-1.3M}} \\
        \midrule
    
        \textbf{XR-Transf. $\left(\vW\right)$} & 50.14 & 39.98 & 20.06 & 27.79 \\
        \textbf{CascadeXML $\left(\vW\right)$} & 47.82	& 38.31	& 17.17	& 24.76 \\

        \textbf{NGAME $\left(\vW\right)$} & 54.69 & 42.80 & 30.01 & 35.29\\

        \textbf{DEXA $\left(\vW\right)$} & 55.76 & 42.96 & 29.12 & 34.86\\
        
        \textbf{Renée $\left(\vW\right)$~\ding{70}} & 56.04 & 45.32 & 28.54 & 36.14\\

        \midrule
        \textbf{PRIME $\left(\theta\right)$} & 58.58 & 45.44 & \textbf{32.14} & \textbf{39.07}\\
        
        \textbf{PRIME $\left(\vW\right)$} & \textbf{59.62} & \textbf{46.75} & \underline{31.20} & \underline{38.64}\\
        \midrule
        \multicolumn{5}{c}{\textbf{LF-Wikipedia-500K}} \\
        \midrule
        \textbf{CascadeXML$\left(\vW\right)$} & 81.13 & 49.12 & 32.12 & 49.37\\
        \textbf{LightXML$\left(\vW\right)$} & 81.59 & 47.64 & 31.99 &	46.53\\
        \textbf{ELIAS$\left(\vW\right)$} & 81.26	& 48.82 & 35.02 &	51.13\\
        \textbf{PINA $\left(\vW\right)$} & 82.83 & 50.11 & - & -\\
        \textbf{NGAME $\left(\vW\right)$} & 84.01 & 49.97 & 41.25 & 57.04\\
        \textbf{DEXA $\left(\vW\right)$} & 84.92 & 50.51 & \underline{42.59} & 58.33\\
        \textbf{Renée $\left(\vW\right)$~\ding{70}} & 84.95 & \textbf{51.68} & 39.89 & 56.70\\
        
        \textbf{OAK $\left(\theta,\vW\right)$~\ding{71}} & 85.23 & 50.79 & \textbf{45.28} & \textbf{60.80}\\

        \midrule
        \textbf{PRIME $\left(\theta\right)$} & 85.37 & 50.92 & 40.60 & 57.74\\
        \textbf{PRIME $\left(\vW\right)$} & \textbf{85.75} & \underline{51.58} & 40.29 & \underline{58.61}\\
        
        \bottomrule 
        \end{tabular}
}
\caption{Comparative evaluation against XMC methods using OVA classifiers ($\vW$). Note that $\left(\theta,\vW\right)$ methods ensemble the encoder and classifier predictions. \ding{70} denotes brute-force algorithm and \ding{71} algorithm using extra meta-data information. Bold and underlined denote best and second best.}
\label{tab:results_classifiers}
\end{table}

\begin{figure}[h]
      \centering
      \includegraphics[width=0.95\linewidth]{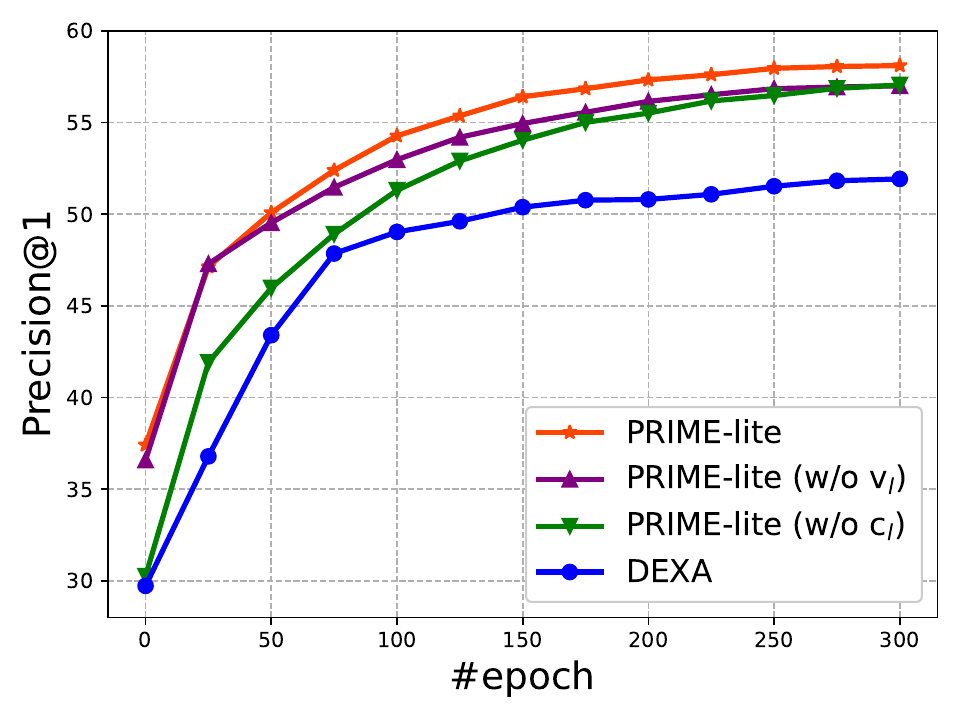}
        \caption{Impact of key components of the Label Prototype Network, \ie centroids ($\vc_l$) and free vectors ($\vv_l$). 
        For convenience, we report results for PRIME-lite, single positive and half batch size variant of PRIME.}
        \label{fig:p1_with_epochs}
\end{figure}

\begin{table} [h]
\renewcommand{\arraystretch}{1.1}
\begin{centering}
\resizebox{0.9\columnwidth}{!}{
\begin{tabular}{l|c|c|c|c}
\toprule
\multirow{2}{*}{\textbf{Method}} & \textbf{Batch} & \textbf{Negative} & \multirow{2}{*}{\textbf{P@1}} & \multirow{2}{*}{\textbf{PSP@5}}\tabularnewline
 & \textbf{size} & \textbf{pool size} &  & \tabularnewline
\midrule 
\multicolumn{5}{c}{\textbf{LF-AmazonTitles-1.3M}}\tabularnewline
\midrule 
\multirow{3}{*}{\textbf{DEXML}} & \multirow{3}{*}{8192} & $\sim$ 1.3M & 58.40 & 36.58\tabularnewline
 &  & $\sim$ 90K & 54.01 & -\tabularnewline
 &  & $\sim$ 16K & 49.16 & -\tabularnewline
\midrule 
\textbf{\alg} & 3200 & $\sim$ 6.4K & 58.58 & 39.07\tabularnewline
\midrule 
\multicolumn{5}{c}{\textbf{LF-Wikipedia-500K}}\tabularnewline
\midrule 
\multirow{3}{*}{\textbf{DEXML}} & \multirow{3}{*}{2048} & $\sim$ 501K & 85.78 & 58.97\tabularnewline
&  & $\sim$ 22K & 84.77 & -\tabularnewline
 &  & $\sim$ 4K & 82.85 & -\tabularnewline
\midrule 
\textbf{PRIME} & 512  & $\sim$ 1K & 85.37 & 57.74\tabularnewline
\bottomrule 
\end{tabular}
}
\par\end{centering}
\caption{\label{tab:PRIMEvsDEXML}Comparison of PRIME and DEXML demonstrating the superiority of PRIME under similar budget.}

\end{table}
\begin{table}[ht!]
\centering
\setlength{\tabcolsep}{5pt}
\resizebox{1.0\linewidth}{!}{
  \begin{tabular}{@{}l|cccc@{}}
  \toprule
  \textbf{\alg-lite variations} & \textbf{P@1} & \textbf{PSP@1} & \textbf{R@100} \\ \midrule
    \textbf{w/o [$\mathcal{L}_T^{cd}(\vh_{q},\vh_{l})$, $m_d$, $\mathcal{L}_T^{cd}(\vh_{l},\vh_{q})$, $\mathcal{R}$]} & 51.71 & 26.27 & 57.02 \\ 
    \textbf{w/o [$m_d$, $\mathcal{L}_T^{cd}(\vh_{l},\vh_{q})$, $\mathcal{R}$]}  & 56.34 & 31.23 & 60.79     \\ 
    \textbf{w/o [$\mathcal{L}_T^{cd}(\vh_{l},\vh_{q})$, $\mathcal{R}$]} & 57.40 & 29.69 & 62.30       \\ 
    w/o [$\mathcal{R}$] & 58.25 & 30.83 & 62.78    \\ 
    \midrule
    \textbf{\alg-lite} & 58.10 & 31.68 & 63.45   \\ 
    \textbf{\alg} & \textbf{58.58} & \textbf{32.14} & \textbf{63.56}   \\
    \bottomrule
\end{tabular}
}\caption{Ablation study in LF-AmazonTitles-1.3M over various components of PRIME's loss function.}
    \label{tab:ablation_loss}
\end{table}

\subsection{PRIME vs brute-force DEXML} \label{subsec:PRIMEvsDEXML}
DEXML~\cite{Gupta24} has demonstrated that, in the absence of classifiers, it is possible to boost XMC performance when using the brute-force approach of pairing every query with all negative labels in the dataset. However, this setup requires substantially more resources than the rest of the algorithms compared in Subsection~\ref{subsec:ResultsComparison}. 

In Table~\ref{tab:PRIMEvsDEXML}, we show that \alg outperforms DEXML for all relaxed setups, while requiring substantially less resources. Notably, \alg achieves better (Amazon data) or competitive (Wikipedia data) performance than the most computationally demanding configuration using $\sim$$2.5$ orders of magnitude more negatives and $2.5 - 4\times$ bigger batch sizes.

Furthermore, as we can observe from Table~\ref{tab:ablation_loss} (\alg-lite vs \alg) it is reasonable to think that increasing PRIME's batch size and number of positives would improve performance, while still using much less resources than DEXML. However, our focus in this work is to propose a high performing method that can run in a single GPU, aiming at keeping the scaling capabilities needed in XMC.

\subsection{Understanding PRIME's components}

In this section we analyze key components of our Label Prototype Network and the loss function. We run a light configuration of PRIME (lite) for faster experimentation (one positive and half batch size).

Figure~\ref{fig:p1_with_epochs} depicts the performance of PRIME-lite compared to variations when ablating free vectors (w/o $v_l$) and centroids (w/o $c_l$). While using free vectors clearly boosts final performance, the addition of centroids have a remarkable impact on faster convergence reaching DEXA's final performance $5\times$ faster.

We further analyze the impact of the number of free vectors in Appendix~\ref{app:free_vectors}. As expected, we observe a positive trend in performance when we increase the number of free vectors.

We analyze in Table~\ref{tab:ablation_loss} the impact of the different terms in \alg's loss function (Equation~\ref{Eq:FinalLoss}).

Results obtained demonstrate that: 1) although we use label prototypes at inference time, optimizing text-based query-to-label and label-to-query relations is crucial for higher encoder quality; and 2) adopting the proposed dynamic margin $m_d$ and the regularization consistently improve the results. Finally, increasing the batch size and adding one positive during training achieves best performance.

\section{Conclusion}
\label{sec5:conclusion}
This paper proposes \alg, a novel prototypical extreme multi-label classification method demonstrating that encoder-based models do not need to sacrifice efficiency to deliver state-of-the-art performance. \alg's key contributions revolves around two key concepts, leveraging dynamic margin-based contrastive learning and label prototypes. 
The proposed Label Prototype Network learns to efficiently aggregate information from text-based embeddings, label centroids and learnable free vectors resulting in highly informative label prototypes for improved results and faster convergence to high performances. 
Additionally, by equipping our multi-objective contrastive optimization with \alg's novel dynamic margin loss, we demonstrate better adaptability to the high granularity and ambiguity posed by extreme label spaces. Our experiments show that \alg outperforms previous works and the leading brute-force approaches in all experiments and for most metrics.
\section*{Limitations}
\label{sec5:limitations}

Our PRIME approach is based on encoder-only architectures, however, decoder-only models have recently demonstrated to be powerful embedding extractors~\cite{Muennighoff24, Wang24} and they might boost XMC performance. However, XMC revolves around efficiency and, even if this decoder-only methods can benefit from efficient fine-tuning, both training and inference time would require higher computational demands.

Furthermore, while PRIME demonstrates top performance on standard XMC baselines, testing it on larger datasets would be interesting to understand scaling limitations when going beyond 1.3M labels, \eg how to define the size of the free vector bank. Additionally, as most related work do, we overlook dealing with unseen labels and we see it as a limitation that future work should address.
\section*{Acknowledgements}
The authors want to thank Alejandro Barón for the fruitful discussions during the development of this work.
Grateful for the invaluable guidance, inspiration, and friendship of Dr. Kevin McGuinness, Diego Ortego wants to honour his legacy. His memory continues to inspire those who were fortunate to know him. Kevin, we miss you.

\bibliography{references}

\begin{thebibliography}{55}
\providecommand{\natexlab}[1]{#1}

\bibitem[{Babbar and Sch\"{o}lkopf(2017)}]{Babbar17}
R.~Babbar and B.~Sch\"{o}lkopf. 2017.
\newblock {DiSMEC: Distributed Sparse Machines for Extreme Multi-label
  Classification}.
\newblock In \emph{ACM International Conference on Web Search and Data Mining
  (WSDM)}.

\bibitem[{Babbar and Sch\"{o}lkopf(2019)}]{Babbar19}
R.~Babbar and B.~Sch\"{o}lkopf. 2019.
\newblock {Data scarcity, robustness and extreme multi-label classification}.
\newblock \emph{Machine Learning (ML)}.

\bibitem[{Bhatia et~al.(2016)Bhatia, Dahiya, Jain, Kar, Mittal, Prabhu, and
  Varma}]{XMLRepo}
K.~Bhatia, K.~Dahiya, H.~Jain, P.~Kar, A.~Mittal, Y.~Prabhu, and M.~Varma.
  2016.
\newblock \href {http://manikvarma.org/downloads/XC/XMLRepository.html} {{The
  Extreme Classification Repository: Multi-label Datasets \& Code}}.

\bibitem[{Bhatia et~al.(2015)Bhatia, Jain, Kar, Varma, and Jain}]{Bhatia15}
K.~Bhatia, H.~Jain, P.~Kar, M.~Varma, and P.~Jain. 2015.
\newblock {Sparse Local Embeddings for Extreme Multi-label Classification}.
\newblock In \emph{Advances in Neural Information Processing Systems
  (NeurIPS)}.

\bibitem[{Chien et~al.(2023)Chien, Zhang, Jiang, Chang, Milenkovic, and
  Yu}]{Chien23}
E.~Chien, C.-J. Zhang, J.~Hsieh, J.-Y. Jiang, W.-C. Chang, O.~Milenkovic, and
  H.-F. Yu. 2023.
\newblock Pina: Leveraging side information in extreme multi-label
  classification via predicted instance neighborhood aggregation.
\newblock In \emph{International Conference on Machine Learning (ICML)}.

\bibitem[{Chopra et~al.(2005)Chopra, Hadsell, and LeCun}]{LeCun05}
S.~Chopra, R.~Hadsell, and Y.~LeCun. 2005.
\newblock {Learning a similarity metric discriminatively, with application to
  face verification}.
\newblock In \emph{IEEE Computer Society Conference on Computer Vision and
  Pattern Recognition (CVPR)}.

\bibitem[{Dahiya et~al.(2021{\natexlab{a}})Dahiya, Agarwal, Saini, Gururaj,
  Jiao, Singh, Agarwal, Kar, and Varma}]{Dahiya21b}
K.~Dahiya, A.~Agarwal, D.~Saini, K.~Gururaj, J.~Jiao, A.~Singh, S.~Agarwal,
  P.~Kar, and M.~Varma. 2021{\natexlab{a}}.
\newblock {SiameseXML: Siamese Networks meet Extreme Classifiers with 100M
  Labels}.
\newblock In \emph{International Conference on Machine Learning (ICML)}.

\bibitem[{Dahiya et~al.(2023{\natexlab{a}})Dahiya, Gupta, Saini, Soni, Wang,
  Dave, Jiao, Gururaj, Dey, Singh, Hada, Jain, Paliwal, Mittal, Mehta, Ramjee,
  Agarwal, Kar, and Varma}]{Dahiya23}
K.~Dahiya, N.~Gupta, D.~Saini, A.~Soni, Y.~Wang, K.~Dave, J.~Jiao, K.~Gururaj,
  P.~Dey, A.~Singh, D.~Hada, V.~Jain, B.~Paliwal, A.~Mittal, S.~Mehta,
  R.~Ramjee, S.~Agarwal, P.~Kar, and M.~Varma. 2023{\natexlab{a}}.
\newblock Ngame: Negative mining-aware mini-batching for extreme
  classification.
\newblock In \emph{ACM International Conference on Web Search and Data Mining
  (WSDM)}.

\bibitem[{Dahiya et~al.(2021{\natexlab{b}})Dahiya, Saini, Mittal, Shaw, Dave,
  Soni, Jain, Agarwal, and Varma}]{Dahiya21}
K.~Dahiya, D.~Saini, A.~Mittal, A.~Shaw, K.~Dave, A.~Soni, H.~Jain, S.~Agarwal,
  and M.~Varma. 2021{\natexlab{b}}.
\newblock {DeepXML: A Deep Extreme Multi-Label Learning Framework Applied to
  Short Text Documents}.
\newblock In \emph{ACM International Conference on Web Search and Data Mining
  (WSDM)}.

\bibitem[{Dahiya et~al.(2023{\natexlab{b}})Dahiya, Yadav, Sondhi, Saini, Mehta,
  Jiao, Agarwal, Kar, and Varma}]{Dahiya23b}
K.~Dahiya, S.~Yadav, S.~Sondhi, D.~Saini, S.~Mehta, J.~Jiao, S.~Agarwal,
  P.~Kar, and M.~Varma. 2023{\natexlab{b}}.
\newblock Deep encoders with auxiliary parameters for extreme classification.
\newblock In \emph{ACM SIGKDD Conference on Knowledge Discovery and Data Mining
  (KDD)}.

\bibitem[{Dopierre et~al.(2021)Dopierre, Gravier, and Logerais}]{Dopierre21}
T.~Dopierre, C.~Gravier, and W.~Logerais. 2021.
\newblock {ProtAugment: Unsupervised diverse short-texts paraphrasing for
  intent detection meta-learning}.
\newblock In \emph{Annual Meeting of the Association for Computational
  Linguistics (ACL)}.

\bibitem[{Gao et~al.(2021)Gao, Yao, and Chen}]{Gao21}
T.~Gao, X.~Yao, and D.~Chen. 2021.
\newblock {SimCSE: Simple Contrastive Learning of Sentence Embeddings}.
\newblock In \emph{Conference on Empirical Methods in Natural Language
  Processing (EMNLP)}.

\bibitem[{Gupta et~al.(2022)Gupta, Chen, Yu, Hsieh, and Dhillon}]{Gupta22}
N.~Gupta, P.~H. Chen, H.-F. Yu, Cho-J. Hsieh, and I.~S. Dhillon. 2022.
\newblock Elias: End-to-end learning to index and search in large output
  spaces.
\newblock In \emph{Advances in Neural Information Processing Systems
  (NeurIPS)}.

\bibitem[{Gupta et~al.(2024)Gupta, Khatri, Rawat, Bhojanapalli, Jain, and
  Dhillon}]{Gupta24}
N.~Gupta, D.~Khatri, A.S. Rawat, S.~Bhojanapalli, P.~Jain, and I.~Dhillon.
  2024.
\newblock {Dual-Encoders for Extreme Multi-Label Classification}.
\newblock In \emph{International Conference on Learning Representations
  (ICLR)}.

\bibitem[{Jain et~al.(2019)Jain, Balasubramanian, Chunduri, and Varma}]{Jain19}
H.~Jain, V.~Balasubramanian, B.~Chunduri, and M.~Varma. 2019.
\newblock {Slice: Scalable Linear Extreme Classifiers trained on 100 Million
  Labels for Related Searches}.
\newblock In \emph{ACM International Conference on Web Search and Data Mining
  (WSDM)}.

\bibitem[{Jain et~al.(2016)Jain, Prabhu, and Varma}]{Jain16}
H.~Jain, Y.~Prabhu, and M.~Varma. 2016.
\newblock {Extreme Multi-label Loss Functions for Recommendation, Tagging,
  Ranking and Other Missing Label Applications}.
\newblock In \emph{ACM SIGKDD Conference on Knowledge Discovery and Data Mining
  (KDD)}.

\bibitem[{Jain et~al.(2023)Jain, Prakash, Saini, Jiao, Ramjee, and
  Varma}]{Jain23}
V.~Jain, J.~Prakash, D.~Saini, J.~Jiao, R.~Ramjee, and M.~Varma. 2023.
\newblock Renée: End-to-end training of extreme classification models.
\newblock In \emph{Conference on Machine Learning and Systems (MLSys)}.

\bibitem[{Jiang et~al.(2021)Jiang, Wang, Sun, Yang, Zhao, and Zhuang}]{Jiang21}
T.~Jiang, D.~Wang, L.~Sun, H.~Yang, Z.~Zhao, and F.~Zhuang. 2021.
\newblock {LightXML: Transformer with Dynamic Negative Sampling for
  High-Performance Extreme Multi-label Text Classification}.
\newblock In \emph{Association for the Advancement of Artificial Intelligence
  Conference on Artificial Intelligence (AAAI)}.

\bibitem[{Johnson et~al.(2019)Johnson, Douze, and J{\'e}gou}]{Johnson19}
J.~Johnson, M.~Douze, and H.~J{\'e}gou. 2019.
\newblock Billion-scale similarity search with {GPUs}.
\newblock \emph{IEEE Transactions on Big Data}.

\bibitem[{Karpukhin et~al.(2020)Karpukhin, Oguz, Min, Lewis, Wu, Edunov, Chen,
  and Yih}]{Karpukhin20}
V.~Karpukhin, B.~Oguz, S.~Min, P.~Lewis, L.~Wu, S.~Edunov, D.~Chen, and W.-T.
  Yih. 2020.
\newblock Dense passage retrieval for open-domain question answering.
\newblock In \emph{Conference on Empirical Methods in Natural Language
  Processing (EMNLP)}.

\bibitem[{Kharbanda et~al.(2022)Kharbanda, Banerjee, Schultheis, and
  Babbar}]{Kharbanda22}
S.~Kharbanda, A.~Banerjee, E.~Schultheis, and R.~Babbar. 2022.
\newblock Cascadexml: Rethinking transformers for end-to-end multi-resolution
  training in extreme multi-label classification.
\newblock In \emph{Advances in Neural Information Processing Systems
  (NeurIPS)}.

\bibitem[{Kim et~al.(2020)Kim, Kim, Cho, and Kwak}]{Kim20}
S.~Kim, D.~Kim, M.~Cho, and S.~Kwak. 2020.
\newblock {Proxy Anchor Loss for Deep Metric Learning}.
\newblock In \emph{IEEE/CVF Computer Vision and Pattern Recognition Conference
  (CVPR)}.

\bibitem[{Kong et~al.(2022)Kong, Khadanga, Li, Gupta, Zhang, Xu, and
  Bendersky}]{Kong22}
W.~Kong, S.~Khadanga, C.~Li, S.~Gupta, M.~Zhang, W.~Xu, and M.~Bendersky. 2022.
\newblock Multi-aspect dense retrieval.
\newblock In \emph{ACM SIGKDD Conference on Knowledge Discovery and Data Mining
  (KDD)}.

\bibitem[{Li et~al.(2021)Li, Liu, Xiong, and Liu}]{Li21}
Y.~Li, Z.~Liu, C.~Xiong, and Z.~Liu. 2021.
\newblock {More Robust Dense Retrieval with Contrastive Dual Learning}.
\newblock In \emph{ACM SIGIR International Conference on Theory of Information
  Retrieval (ICTIR)}.

\bibitem[{Liu et~al.(2021)Liu, Xu, Wen, Jiang, Wu, and Yuan}]{Liu21}
Z.~Liu, H.~Xu, Y.~Wen, N.~Jiang, H.~Wu, and X.~Yuan. 2021.
\newblock {TEMP: Taxonomy Expansion with Dynamic Margin Loss through
  Taxonomy-Paths}.
\newblock In \emph{Conference on Empirical Methods in Natural Language
  Processing (EMNLP)}.

\bibitem[{Malkov and Yashunin(2020)}]{MalkovY16}
A.~Y. Malkov and D.~A. Yashunin. 2020.
\newblock {Efficient and robust approximate nearest neighbor search using
  Hierarchical Navigable Small World graphs}.
\newblock \emph{IEEE Transactions on Pattern Analysis and Machine Intelligence
  (TPAMI)}.

\bibitem[{Mesquita et~al.(2022)Mesquita, Martins, and Almeida}]{Mesquita22}
T.~Mesquita, B.~Martins, and M.~Almeida. 2022.
\newblock {Dense Template Retrieval for Customer Support}.
\newblock In \emph{International Conference on Computational Linguistics
  (COLING)}.

\bibitem[{Mittal et~al.(2022)Mittal, Dahiya, Malani, Ramaswamy, Kuruvilla,
  Ajmera, Chang, Agrawal, Kar, and Varma}]{Mittal22}
A.~Mittal, K.~Dahiya, S.~Malani, J.~Ramaswamy, S.~Kuruvilla, J.~Ajmera,
  K.~Chang, S.~Agrawal, P.~Kar, and M.~Varma. 2022.
\newblock Multimodal extreme classification.
\newblock In \emph{IEEE/CVF Computer Vision and Pattern Recognition Conference
  (CVPR)}.

\bibitem[{Mittal et~al.(2021)Mittal, Sachdeva, Agrawal, Agarwal, Kar, and
  Varma}]{Mittal21b}
A.~Mittal, N.~Sachdeva, S.~Agrawal, S.~Agarwal, P.~Kar, and M.~Varma. 2021.
\newblock {ECLARE: Extreme Classification with Label Graph Correlations}.
\newblock In \emph{International Conference on World Wide Web (WWW)}.

\bibitem[{Mohan et~al.(2024)Mohan, Saini, Mittal, Chowdhury, Paliwal, Jiao,
  Gupta, and Varma}]{Mohan24}
S.~Mohan, D.~Saini, A.~Mittal, S.R. Chowdhury, B.~Paliwal, J.~Jiao, M.~Gupta,
  and M.~Varma. 2024.
\newblock Oak: Enriching document representations using auxiliary knowledge for
  extreme classification.
\newblock In \emph{International Conference on Machine Learning (ICML)}.

\bibitem[{Muennighoff et~al.(2024)Muennighoff, Su, Wang, Yang, Wei, Yu, Singh,
  and Kiela}]{Muennighoff24}
N.~Muennighoff, H.~Su, L.~Wang, N.~Yang, F.~Wei, T.~Yu, A.~Singh, and D.~Kiela.
  2024.
\newblock Generative representational instruction tuning.
\newblock \emph{arXiv preprint arXiv:2402.09906}.

\bibitem[{Nguyen et~al.(2022)Nguyen, Nguyen, and Tiulpin}]{Nguyen22}
K.~Nguyen, H.~H. Nguyen, and A.~Tiulpin. 2022.
\newblock {AdaTriplet: Adaptive Gradient Triplet Loss with Automatic Margin
  Learning for Forensic Medical Image Matching}.
\newblock In \emph{International Conference on Medical Image Computing and
  Computer Assisted Intervention (MICCAI)}.

\bibitem[{Prabhu et~al.(2018{\natexlab{a}})Prabhu, Kag, Gopinath, Dahiya,
  Harsola, Agrawal, and Varma}]{Prabhu18}
Y.~Prabhu, A.~Kag, S.~Gopinath, K.~Dahiya, S.~Harsola, R.~Agrawal, and
  M.~Varma. 2018{\natexlab{a}}.
\newblock {Extreme multi-label learning with label features for warm-start
  tagging, ranking and recommendation}.
\newblock In \emph{ACM International Conference on Web Search and Data Mining
  (WSDM)}.

\bibitem[{Prabhu et~al.(2018{\natexlab{b}})Prabhu, Kag, Harsola, Agrawal, and
  Varma}]{Prabhu18b}
Y.~Prabhu, A.~Kag, S.~Harsola, R.~Agrawal, and M.~Varma. 2018{\natexlab{b}}.
\newblock {Parabel: Partitioned label trees for extreme classification with
  application to dynamic search advertising}.
\newblock In \emph{International Conference on World Wide Web (WWW)}.

\bibitem[{Prabhu and Varma(2014)}]{Prabhu14}
Y.~Prabhu and M.~Varma. 2014.
\newblock {FastXML: A Fast, Accurate and Stable Tree-classifier for eXtreme
  Multi-label Learning}.
\newblock In \emph{ACM SIGKDD Conference on Knowledge Discovery and Data Mining
  (KDD)}.

\bibitem[{Reimers and Gurevych(2019)}]{Reimers19}
N.~Reimers and I.~Gurevych. 2019.
\newblock {Sentence-bert: Sentence embeddings using siamese bert-networks}.
\newblock \emph{Conference on Empirical Methods in Natural Language Processing
  (EMNLP)}.

\bibitem[{Ren et~al.(2024)Ren, Chen, Wang, and Hua}]{Ren24}
L.~Ren, C.~Chen, L.~Wang, and K.~Hua. 2024.
\newblock {Towards Improved Proxy-based Deep Metric Learning via Data-Augmented
  Domain Adaptation}.
\newblock In \emph{Association for the Advancement of Artificial Intelligence
  Conference on Artificial Intelligence (AAAI)}.

\bibitem[{Saini et~al.(2021)Saini, Jain, Dave, Jiao, Singh, Zhang, and
  Varma}]{Saini21}
D.~Saini, A.K. Jain, K.~Dave, J.~Jiao, A.~Singh, R.~Zhang, and M.~Varma. 2021.
\newblock {GalaXC: Graph Neural Networks with Labelwise Attention for Extreme
  Classification}.
\newblock In \emph{International Conference on World Wide Web (WWW)}.

\bibitem[{Schroff et~al.(2015)Schroff, Kalenichenko, and Philbin}]{Schroff15}
F.~Schroff, D.~Kalenichenko, and J.~Philbin. 2015.
\newblock {Facenet: A unified embedding for face recognition and clustering}.
\newblock In \emph{IEEE/CVF Computer Vision and Pattern Recognition Conference
  (CVPR)}.

\bibitem[{Snell et~al.(2017)Snell, Swersky, and Zemel}]{Snell17}
J.~Snell, K.~Swersky, and R.S. Zemel. 2017.
\newblock {Prototypical Networks for Few-shot Learning}.
\newblock In \emph{Advances in Neural Information Processing Systems
  (NeurIPS)}.

\bibitem[{So et~al.(2023)So, Lim, Kim, Oh, and Song}]{So23}
J.~So, Y.~Lim, Y.~Kim, C.~Oh, and K.~Song. 2023.
\newblock Robust contrastive learning with dynamic mixed margin.
\newblock \emph{IEEE Access}.

\bibitem[{Song et~al.(2022)Song, Huang, Xue, and Hu}]{Song22}
X.~Song, L.~Huang, H.~Xue, and S.~Hu. 2022.
\newblock {Supervised Prototypical Contrastive Learning for Emotion Recognition
  in Conversation}.
\newblock In \emph{Conference on Empirical Methods in Natural Language
  Processing (EMNLP)}.

\bibitem[{Sun et~al.(2024)Sun, Bi, Guo, Yang, Zhang, Liu, Zhang, and
  Cheng}]{Sun24}
X.~Sun, K.~Bi, J.~Guo, S.~Yang, Q.~Zhang, Z.~Liu, G.~Zhang, and X.~Cheng. 2024.
\newblock A multi-granularity-aware aspect learning model for multi-aspect
  dense retrieval.
\newblock In \emph{ACM International Conference on Web Search and Data Mining
  (WSDM)}.

\bibitem[{Tagami(2017)}]{Tagami17}
Y.~Tagami. 2017.
\newblock {AnnexML: Approximate Nearest Neighbor Search for Extreme Multi-label
  Classification}.
\newblock In \emph{ACM SIGKDD Conference on Knowledge Discovery and Data Mining
  (KDD)}.

\bibitem[{Van~den Oord et~al.(2018)Van~den Oord, Li, and Vinyals}]{Vinyals18}
A.~Van~den Oord, Y.~Li, and O.~Vinyals. 2018.
\newblock Representation learning with contrastive predictive coding.
\newblock \emph{arXiv preprint arXiv:1807.03748}.

\bibitem[{Vaswani et~al.(2017)Vaswani, Shazeer, Parmar, Uszkoreit, Jones,
  Gomez, Kaiser, and Polosukhin}]{Vaswani17}
A.~Vaswani, N.~Shazeer, N.~Parmar, J.~Uszkoreit, L.~Jones, A.N. Gomez, Ł.
  Kaiser, and I.~Polosukhin. 2017.
\newblock {Attention Is All You Need}.
\newblock In \emph{Advances in Neural Information Processing Systems
  (NeurIPS)}.

\bibitem[{Wang et~al.(2022)Wang, Yang, Huang, Jiao, Yang, Jiang, Majumder, and
  Wei}]{Wang22}
L.~Wang, N.~Yang, X.~Huang, B.~Jiao, L.~Yang, D.~Jiang, R.~Majumder, and
  F.~Wei. 2022.
\newblock Text embeddings by weakly-supervised contrastive pre-training.
\newblock \emph{arXiv preprint arXiv:2212.03533}.

\bibitem[{Wang et~al.(2024)Wang, Yang, Huang, Yang, Majumder, and Wei}]{Wang24}
L.~Wang, N.~Yang, X.~Huang, L.~Yang, R.~Majumder, and F.~Wei. 2024.
\newblock {Improving Text Embeddings with Large Language Models}.
\newblock In \emph{Annual Meeting of the Association for Computational
  Linguistics (ACL)}.

\bibitem[{Xiong et~al.(2021)Xiong, Xiong, Li, Tang, Liu, Bennett, Ahmed, and
  Overwijk}]{Xiong20}
L.~Xiong, C.~Xiong, Y.~Li, K.-F. Tang, J.~Liu, P.~Bennett, J.~Ahmed, and
  A.~Overwijk. 2021.
\newblock {Approximate nearest neighbor negative contrastive learning for dense
  text retrieval}.
\newblock In \emph{International Conference on Learning Representations
  (ICLR)}.

\bibitem[{Yang et~al.(2021)Yang, Liu, Xiao, Li, Lian, Agrawal, Singh, Sun, and
  Xie}]{Yang21}
J.~Yang, Z.~Liu, S.~Xiao, C.~Li, D.~Lian, S.~Agrawal, A.~Singh, G.~Sun, and
  X.~Xie. 2021.
\newblock Graphformers: Gnn-nested transformers for representation learning on
  textual graph.
\newblock In \emph{Advances in Neural Information Processing Systems
  (NeurIPS)}.

\bibitem[{Yen et~al.(2017)Yen, Huang, Dai, Ravikumar, Dhillon, and
  Xing}]{Yen17}
E.~H.~I. Yen, X.~Huang, W.~Dai, P.~Ravikumar, I.~Dhillon, and E.~Xing. 2017.
\newblock {PPDSparse: A Parallel Primal-Dual Sparse Method for Extreme
  Classification}.
\newblock In \emph{ACM SIGKDD Conference on Knowledge Discovery and Data Mining
  (KDD)}.

\bibitem[{You et~al.(2019)You, Dai, Zhang, Mamitsuka, and Zhu}]{You18}
R.~You, S.~Dai, Z.~Zhang, H.~Mamitsuka, and S.~Zhu. 2019.
\newblock {AttentionXML: Extreme Multi-Label Text Classification with
  Multi-Label Attention Based Recurrent Neural Networks}.
\newblock In \emph{Advances in Neural Information Processing Systems
  (NeurIPS)}.

\bibitem[{Zeng et~al.(2022)Zeng, Yin, Jiang, Wu, and Cao}]{Zeng22}
J.~Zeng, Y.~Yin, Y.~Jiang, S.~Wu, and Y.~Cao. 2022.
\newblock {Contrastive Learning with Prompt-derived Virtual Semantic Prototypes
  for Unsupervised Sentence Embedding}.
\newblock In \emph{Conference on Empirical Methods in Natural Language
  Processing, Findings Track (EMNLP-F)}.

\bibitem[{Zhang et~al.(2021{\natexlab{a}})Zhang, Li, Xiao, Zhu, Nallapati,
  Arnold, and Xiang}]{Zhang21b}
D.~Zhang, S.-W. Li, W.~Xiao, H.~Zhu, R.~Nallapati, A.O. Arnold, and B.~Xiang.
  2021{\natexlab{a}}.
\newblock {Pairwise Supervised Contrastive Learning of Sentence
  Representations}.
\newblock In \emph{Conference on Empirical Methods in Natural Language
  Processing (EMNLP)}.

\bibitem[{Zhang et~al.(2021{\natexlab{b}})Zhang, Chang, Yu, and
  Dhillon}]{Zhang21}
J.~Zhang, W.~C. Chang, H.~F. Yu, and I.~Dhillon. 2021{\natexlab{b}}.
\newblock Fast multi-resolution transformer fine-tuning for extreme multi-label
  text classification.
\newblock In \emph{Advances in Neural Information Processing Systems
  (NeurIPS)}.

\end{thebibliography}

\clearpage
\onecolumn

\begin{appendices}

\section{Triplet loss with clipped dynamic margin. Propositions and proofs.}
\label{sec:appendix}

Consider $\cL_T$ in Eq.~\ref{eq:app_triplet_ang} be the simplified version of the original triplet loss~\cite{Schroff15} in the angular domain,
\begin{eqnarray}
    \nonumber
    {\cL}_T & = & \max{\left(s_{an} - s_{ap} + m, 0 \right)} \\
    & = & 
    \begin{cases}
        \Delta s_{a}^{n-p} + m, & \textrm{if~~} \Delta s_{a}^{p-n} \le m\tabularnewline
        0, & \textrm{otherwise}
    \end{cases},
    \label{eq:app_triplet_ang}
\end{eqnarray}
where $s_{an}$ and $s_{ap}$ denote the anchor-negative and the anchor-positive cosine similarities, respectively. For convenience, we denote $\Delta s_{a}^{n - p} = s_{an} - s_{ap}$ and $\Delta s_{a}^{p-n} = s_{ap} - s_{an}$ as the differences of the cosine similarities between the tuples anchor-positive and anchor-negative, respectively. Assuming normalized feature vectors, Eq.~\ref{eq:app_triplet_ang} can take the values $m \in [0, 2)$.

The traditional triplet loss formulation imposes a fixed margin during training such that all negatives, regardless the degree of similarity to the positive, are pushed apart from the anchor while positives are pulled close to the anchor given condition $\Delta s_{a}^{p-n}\le m$. The partial derivatives with respect to $s_{ap}$ and $s_{an}$ look as follows:

\begin{equation}
    \left( \frac{\partial {\cL}_T}{\partial s_{ap}}, \frac{\partial {\cL}_T}{\partial s_{an}} \right) = 
    \begin{cases}
        (-1, 1), & \textrm{if~~} \Delta s_{a}^{p-n}\le  m \tabularnewline
        (\;\;\,0, 0), & \textrm{otherwise}
    \end{cases}.
    \label{eq:app_triplet_grad_static}
\end{equation}

\subsection{Proof of proposition 1}
\paragraph{Proposition 1.}\textit{Consider the non-differentiable piece-wise linear function defining the adaptive margin to be ~$m\left(s_{ap}, s_{an}\right) = | s_{ap} - s_{an}|$. Adding $m\left(s_{ap}, s_{an}\right)$ into Eq.~\ref{eq:app_triplet_ang} expands the support of the function by relaxing the margin constraint of the original triplet formulation, inducing a modulation effect that allows semantically similar representations to be projected closer in the embedding space.}

\begin{proof}
Consider $m\left(s_{ap}, s_{an}\right)$ be the non-differentiable piece-wise linear margin function defined as, 

\begin{equation}
    m\left(s_{ap}, s_{an}\right) = 
    \begin{cases}
        s_{ap} - s_{an}, & \textrm{if~~} s_{ap} > s_{an}\tabularnewline
        s_{an} - s_{ap}, & \textrm{if~~} s_{ap} \le s_{an}
    \end{cases},
\end{equation}
and consider as well $(x)_\dagger$ be the operator that detaches $x$ from the computational graph, thus avoiding the gradients to back-propagate. Then, applying the dynamic margin in Eq.~\ref{eq:app_triplet_ang}, the triplet loss becomes:
\begin{equation}
    {\cL}_T^d = 
    \begin{cases}
        s_{an} - s_{ap} + \left( s_{an} - s_{ap}\right)_\dagger, & \textrm{if~~} \Delta s_{a}^{p-n} \le 0 \tabularnewline
        0, & \textrm{otherwise}
    \end{cases}
    \label{eq:app_triplet_dyn}
\end{equation}

In Eq.~\ref{eq:app_triplet_dyn} we show that adding the dynamic margin removes the fixed margin constraint of the original triplet loss formulation enabling the model to learn from any observation that satisfies the inequality $s_{ap} \le s_{an}$ where negatives are closer to the anchor than positives. Besides, we observe that the new function modulates the margin depending on the hardness of the triplets, allowing semantically similar observations ($s_{ap} \approx s_{an}$ with $s_{ap} \le s_{an}$) to be closer in the embedding space. Conversely, semantically different observations are pushed apart with higher strength.

\label{proof1}
\end{proof}

\subsection{Proof of Proposition 2}
\paragraph{Proposition 2.}\textit{
Clipping the values of the dynamic margin partitions the triplet loss landscape allowing positives and negatives in uncertain settings to be projected nearby in the embedding space.
}

\begin{proof}
Consider $\textrm{clip}(x)$ be the clipping function in Eq.~\ref{eq:app_triplet_clip},
\begin{equation}
    \textrm{clip}(x) = 
    \begin{cases}
        x, & \textrm{if}\quad \gamma_{\textrm{min}} \le x \le \gamma_{\textrm{max}} \tabularnewline
        \gamma_{\textrm{min}}, & \textrm{if}\quad x < \gamma_{\textrm{min}} \tabularnewline
        \gamma_{\textrm{max}}, & \textrm{if}\quad x > \gamma_{\textrm{max}}
    \end{cases},
    \label{eq:app_triplet_clip}
\end{equation}
then, the triplet loss with clipped dynamic margin becomes:

\begin{align}
    \nonumber
    \cL_T^{cd} &=
    \begin{cases}
        \max{\left(\Delta s_a^{n-p} + \textrm{clip}(\Delta s_{a}^{p-n})_\dagger, 0\right)}, & \text{if~~} s_{ap} > s_{an} \tabularnewline
        \max{\left(\Delta s_a^{n-p} + \textrm{clip}(\Delta s_{a}^{n-p})_\dagger, 0\right)}, & \text{if~~} s_{ap} \le s_{an}
    \end{cases}\\
    &=
    \begin{cases}
        0, & \textrm{if~~} s_{ap} > s_{an}, ~~\Delta s_a^{p-n} > \gamma_{\textrm{min}} \tabularnewline
        \Delta s_a^{p-n} + \gamma_{min}, & \textrm{if~~} s_{ap} > s_{an},~~ \Delta s_a^{p-n} < \gamma_{\textrm{min}} \tabularnewline
        \Delta s_a^{n-p} + \textrm{clip}\left(\Delta s_{a}^{n-p}\right)_\dagger, & \textrm{if~~} s_{ap} \le s_{an}
    \end{cases},
    \label{eq:app_triplet_clip_dyn}
\end{align}

From Eq.~\ref{eq:app_triplet_clip_dyn} one can easily observe that the norm of the partial derivatives when $s_{ap} \le s_{an}$ might increase by a multiplicative factor of 2 if we do not detach the dynamic margin. We empirically observe that increasing the gradient norm for those cases is preventing the model to converge properly. We argue that doubling the norm of the gradients may generate instabilities during training that impact the inter-class separation and the uniformity of the embedding space. By detaching the dynamic margin from the computational graph, the partial derivatives become:

\begin{equation}
    \left( \frac{\partial {\cL}_T^d}{\partial s_{ap}}, \frac{\partial {\cL}_T^d}{\partial s_{an}} \right) = 
    \begin{cases}
        (\;\;\,0,\;\;\,0), & \textrm{if~~}  s_{ap} > s_{an}, \left(\Delta_a^{p-n} > \gamma_{min} \right)\tabularnewline
        (\;\;\,1, -1), & \textrm{if~~}  s_{ap} > s_{an}, \left(\Delta_a^{p-n} < \gamma_{min} \right) \tabularnewline
        (-1, \;\;\,1), & \textrm{if~~} s_{ap} \le s_{an}
    \end{cases}.
    \label{eq:app_triplet_grad}
\end{equation}

The new formulation introduces a new learning region ($s_{ap} > s_{an},~ \Delta_a^{p-n} < \gamma_{min}$) to the dynamic triplet loss where the gradient direction is inverted, \ie negatives are pulled closer to the anchor while positives are pushed apart until  $s_{ap} \le s_{an}$. We argue that this region, where positives and negatives are nearly equally distant to the anchor, accounts for those cases with a high degree of uncertainty such as ambiguous observations or missing labels, a well known  problem in extreme multi-label settings. By reverting the learning process in this small margin, the network is capable of handling these scenarios.

In summary, we can observe the following cases depending on the hardness of the triplet construction:
\begin{itemize}
    \item \emph{Easy cases $\rightarrow  s_{ap} > s_{an}, ~\Delta_a^{p-n} > \gamma_{min}$}. Positives are closer to the anchor than negatives and positives are reasonably separated from negatives. Similarly to $\cL_T$, we do not back-propagate those observations.
    \item \emph{Hard cases $\rightarrow  s_{ap} \le s_{an}$}. We keep the same gradients (norm and direction) as the original triplet loss for hard cases where negatives are closer to the anchor than positives while keeping the modulation effect described in Proposition 1.
    \item \emph{Ambiguity and missing labels $\rightarrow  s_{ap} > s_{an}, ~\Delta_a^{p-n} < \gamma_{min} $}. The proposed dynamic margin relaxes the original $\cL_T$ by allowing positives and negatives to live in a region where both representations are closer to each other. Indeed, the direction of the partial derivatives are inverted, allowing an inversion of the natural order of positives and negatives with respect to the anchor that accounts for high uncertain observations.
\end{itemize}
\end{proof}
\clearpage
\section{Datasets details} 
\label{app:datasets}

In Table~\ref{tab:DatasetDetails} we present the details of the datasets used in Section~\ref{sec4:experiments}. 
The datasets consider multiple real-world applications. In particular, LF-Wikipedia-500K, and LF-Wikipedia-500K involve predicting the relevant categories given the title or the full Wikipedia page, respectively. The LF-WikiSeeAlso-320K dataset addresses the prediction of similar Wikipedia articles.  On the other hand, LF-AmazonTitles-131K and LF-AmazonTitles-1.3M involve recommending similar products using the product title. The LF-Amazon-131K dataset is similar to LF-AmazonTitles-131K, but adding long product descriptions to the product titles. Finally, the label is endowed with a short description for all of these datastes.

\begin{table*}[h!]
\begin{centering}
\resizebox{0.65\columnwidth}{!}{
\begin{tabular}{l|ccccc}
\midrule

\textbf{Dataset}  & \textbf{\#q train}  & \textbf{\#l} & \textbf{\#q test} & \textbf{\#q/l} & \textbf{\#l/q}\tabularnewline
\midrule

\textbf{LF-AmazonTitles-131K} & 294.8K & 131K & 134.8K & 5.15 & 2.29\tabularnewline

\textbf{LF-Amazon-131K} & 294.8K & 131K & 134.8K & 5.15 & 2.29\tabularnewline

\textbf{LF-AmazonTitles-1.3M} & 2.2M & 1.3M & 0.97M & 38.24 & 22.20\tabularnewline

\textbf{LF-Wikipedia-500K} & 1.8M & 0.5M & 0.8M & 24.75 & 4.77\tabularnewline

\textbf{LF-WikiTitles-500K} & 1.8M & 0.5M & 0.8M & 17.15 & 4.74\tabularnewline

\textbf{LF-WikiSeeAlso-320K} & 693.1K & 312.3K & 177.5K & 2.11 & 4.68\tabularnewline
\midrule
\end{tabular}
}
\par\end{centering}
\caption{\label{tab:DatasetDetails}Dataset statistics for benchmark datasets. Key: \#q (number of queries), \#l (number of labels), \#q/l (number of queries per label), \#l/q (number of labels per query).}

\end{table*}

\section{Comparison against methods using classifiers}
\label{app:app_ova}

Many works in the XMC literature design efficient strategies to train a One-vs-All (OVA) classifier $\vW$. On the one hand, some methods just train the classifier and do not focus on training backbone encoders that can be used to compute predictions, as it is the case for XR-Transformer \cite{Zhang21}, LightXML~\cite{Jiang21}, ELIAS~\cite{Gupta22}, CascadeXML~\cite{Kharbanda22}, PINA~\cite{Chien23} and Renée~\cite{Jain23}. On the other hand, methods such as NGAME~\cite{Dahiya23}, DEXA~\cite{Dahiya23b} and OAK~\cite{Mohan24} train the OVA classifiers after encoder training and, then, ensemble the scores predicted by each of the stages, which increases inference complexity (two ANN search structures are required). Note that OAK also introduces external information to enrich the queries, e.g. hyper-link information in Wikipedia datasets. 
The results presented in Table~\ref{tab:ResultsPRIMEOVAapp} demonstrate that \alg with and without extreme classifiers is competitive and sometimes provides better performance than classifier-based methods. \alg achieves top P@1 in 3 out of 6 datasets (4 if we exclude OAK, which exploits additional meta-data information).

\begin{table*}[ht]
\renewcommand{\arraystretch}{1.0}
\setlength{\tabcolsep}{3.0pt}
\begin{centering}
\resizebox{\textwidth}{!}{
\begin{tabular}{@{}l|cccc|cccc|cccc}
\toprule
\textbf{Method}  & \textbf{P@1} & \textbf{P@5} & \textbf{PSP@1} & \textbf{PSP@5} & \textbf{P@1} & \textbf{P@5} & \textbf{PSP@1} & \textbf{PSP@5} & \textbf{P@1} & \textbf{P@5} & \textbf{PSP@1} & \textbf{PSP@5} \\
\midrule
\textbf{Datasets} $\longrightarrow$ & \multicolumn{4}{c|}{\textbf{LF-Amazon-131K}} & \multicolumn{4}{c|}{\textbf{LF-AmazonTitles-1.3M}} & \multicolumn{4}{c}{\textbf{LF-AmazonTitles-131K}} \tabularnewline
\midrule 

\textbf{XR-Tranf. $\left(\vW\right)$} & 45.61 & 22.32 & 34.93 & 49.24 & 50.14 & 39.98 & 20.06 & 27.79 & 38.10 & 18.32 & 28.86 & 39.59\tabularnewline

\textbf{ELIAS $\left(\vW\right)$} & - & - & - & - & - & - & - & - & 40.13 & 19.54 & 31.05 & 42.88\tabularnewline

\textbf{CascadeXML $\left(\vW\right)$} & - & - & - & - & 47.82 & 38.31 & 17.17 & 24.76 & 35.96 & 18.15 & - & -\tabularnewline

\textbf{NGAME $\left(\vW\right)$} & 46.53 & 22.02 & 38.53 & 50.45 & 54.69 & 42.80 & 28.23 & 34.48 & 44.95 & 21.20 & 38.25 & 48.42\tabularnewline

\textbf{NGAME $\left(\theta,\vW\right)$} & 46.65 & 22.03 & 38.67 & 50.12 & 56.75 & 44.09 & 29.18 & 35.36 & 46.01 & 21.47 & 38.81 & 49.43\tabularnewline

\textbf{DEXA $\left(\vW\right)$} & 47.12 & 22.35 & 38.86 & 50.59 & 55.76 & 42.95 & 30.01 & 35.29 & 45.78 & 21.29 & 38.57 & 48.56\tabularnewline

\textbf{DEXA $\left(\theta,\vW\right)$} & 47.16 & 22.42 & 38.70 & 50.97 & 56.63 & 43.90 & 29.12 & 34.86 & \textbf{46.42} & \underline{21.59} & 39.11 & 49.65\tabularnewline

\textbf{PINA $\left(\vW\right)$} & 46.76 & 23.20 & - & - & - & - & - & - & - & - & - & -\tabularnewline

\textbf{Renée $\left(\vW\right)$~\ding{70}} & 48.05 & 23.26 & 40.11 & \textbf{53.67} & 56.04 & 45.32 & 28.54 & 36.14 & \underline{46.05} & \textbf{22.04} & 39.08 & \textbf{50.48}\tabularnewline

\midrule
\textbf{PRIME $\left(\theta\right)$} & \underline{48.09} & \textbf{23.34} & \textbf{40.48} & \underline{53.43} & \underline{58.58} & \underline{45.44} & \textbf{32.14} & \textbf{39.07} & 44.87 & 21.53 & \textbf{39.59} & \underline{49.73}\tabularnewline

\textbf{PRIME $\left(\vW\right)$} & \textbf{48.20} & \underline{23.28} & \underline{40.16} & 53.22 & \textbf{59.62} & \textbf{46.75} & \underline{31.20} & \underline{38.64} & 45.26 & 21.48 & \underline{39.29} & 49.44\tabularnewline
\midrule
\midrule 
\textbf{Datasets} $\longrightarrow$ & \multicolumn{4}{c|}{\textbf{LF-WikiTitles-500K}} & \multicolumn{4}{c|}{\textbf{LF-Wikipedia-500K}} & \multicolumn{4}{c}{\textbf{LF-WikiSeeAlso-320K}} \tabularnewline
\midrule

\textbf{LightXML$\left(\vW\right)$} & - & - & - & - & 81.57 & 47.64 & 31.99 & 46.53 & 34.50 & 16.83 & 17.85 & 24.16\tabularnewline

\textbf{XR-Transf.$\left(\vW\right)$} & - & - & - & - & 81.62 & 47.85 & 33.58 & 47.81 & 42.57 & 21.30 & 25.18 & 33.79\tabularnewline

\textbf{CascadeXML$\left(\vW\right)$} & \underline{47.29} & \textbf{19.00} & 19.19 & 19.75 & 81.13 & 49.12 & 32.12 & 49.37 & - & - & - & -\tabularnewline

\textbf{NGAME $\left(\vW\right)$} & - & - & - & - & 84.01 & 49.97 & 41.25 & 57.04 & 45.72 & 22.06 & - & -\tabularnewline

\textbf{NGAME $\left(\theta,\vW\right)$} & 39.04 & 16.08	& 23.12	& 23.03 & 84.01 & 49.97 & 41.25 & 57.04 & 47.65 & 23.68 & \underline{33.83} & \textbf{41.03}\tabularnewline

\textbf{DEXA $\left(\vW\right)$} & - & - & - & - & 84.92 & 50.51 & \underline{42.59} & 58.33 & - & - & - & -\tabularnewline

\textbf{DEXA $\left(\theta,\vW\right)$} & \textbf{47.41} & 17.62 & 25.27 & 24.03 & 83.52 & 50.85 & 42.15 & 57.38 & 47.11 & 22.71 & 31.82 & 38.78\tabularnewline

\textbf{Renée $\left(\vW\right)$~\ding{70}} & - & - & - & - & 84.95 & \textbf{51.68} & 39.89 & 56.70 & 47.86 & \textbf{24.05} & 32.02 & \underline{40.90}\tabularnewline
\textbf{PINA $\left(\vW\right)$} & - & - & - & - & 82.83 & 50.11  & - & - & 44.54 & 22.92 & - & -\tabularnewline
\textbf{OAK $\left(\theta,\vW\right)$~\ding{71}} & 44.82 & 17.67 & \textbf{25.79} & \textbf{24.90} & 85.23 & 50.79 & \textbf{45.28} & \textbf{60.80} & \textbf{48.57} & 23.28 & \textbf{33.92} & 40.44\tabularnewline

\midrule
\textbf{PRIME $\left(\theta\right)$} & 46.33 & 18.09 & \underline{25.36} & 24.03 & \underline{85.37} & 50.92 & 40.60 & 57.74 & 48.00 & 23.53 & 32.88 & 40.20\tabularnewline
\textbf{PRIME $\left(\vW\right)$} & 46.75 & \underline{18.44} & 25.13 & \underline{24.31} & \textbf{85.75} & \underline{51.58} & 40.29 & \underline{58.61} & \underline{48.14} & \underline{23.61} & 32.80 & 40.24\tabularnewline

\bottomrule 
\end{tabular}
}

\par\end{centering}
\caption{\label{tab:ResultsPRIMEOVAapp}Comparative evaluation against XMC methods using OVA classifiers ($\vW$). Note that $\left(\theta,\vW\right)$ methods ensemble the encoder and classifier predictions. \ding{70} denotes brute-force algorithm and \ding{71} algorithm using extra meta-data information. Bold and underlined denote best and second best.}
\end{table*}

\clearpage

\section{Effect of the number of free vectors}
\label{app:free_vectors}
We analyze the impact of the number of free vectors used in Table \ref{tab:ablation_n_auxvectors}. More free vectors exhibit better performance, leading to around 1\% boost in P@1. Although this behaviour is aligned with observations made in DEXA~\cite{Dahiya23b} work, free vectors are less important for boosting performance for our method.
\begin{table}
\centering
\resizebox{0.35\linewidth}{!}{
  \begin{tabular}{@{}l|cccc@{}}
  \toprule
   \textbf{\# Auxiliary} & \multirow{2}{*}{\textbf{P@1}} & \multirow{2}{*}{\textbf{P@3}} & \multirow{2}{*}{\textbf{P@5}} \tabularnewline
   \textbf{Vectors} & & & \\ \midrule
    - & 57.00 & 49.67 & 44.50 \\ 
    1,024  & 57.09 & 49.70 & 44.51    \\ 
    4,096 & 57.13 & 49.75 & 44.53      \\ 
    16,384 & 57.47 & 49.97 & 44.73    \\ 
    32768 & 57.63 & 50.09 & 44.83    \\ 
    65,536 & 57.87 & 50.31 & 45.03    \\ 
    131,072 & \textbf{58.10} & \textbf{50.39} & \textbf{45.05}   \\ 
    \bottomrule
\end{tabular}
}\caption{Performance in LF-AmazonTitles-1.3M when varying the number of free vectors used in PRIME-lite.}
    \label{tab:ablation_n_auxvectors}
\end{table}

\section{Complexity calculations} \label{app:complexity}
An encoder-based method will need to compute the embeddings of the queries and labels at every mini-batch.
Consider $B$ and $\cS$ be the batch size and the subset of labels considered in a mini-batch, respectively. Moreover, $K$ is the complexity of computing the embedding of a single query or label item and $L$ is the total number of labels. The complexity of different components is defined as follows: 
\begin{enumerate}
    \item $\bigO{KB}$ for embedding $B$ queries.
    \item  $\bigO{K|\cS|}$ for embedding the labels considered in the mini-batch.
    \item  $\bigO{Bd|\cS|}$ for computing the loss. Recall that $d$ refers to the dimensionality of the final embeddings computed using the encoder.
\end{enumerate}
This results in $\bigO{BK + KL + BdL} = \bigO{KL + BdL}$ for a brute-force approach as $\cS = L$ and $L \gg B$. On the other hand, it translates to $\bigO{BK + B^2d}$ as $ \cS \approx B$ for PRIME as it uses a small pool of labels.

\section{Additional related works} \label{app:related}
Early XMC methods focused on efficient learning of classifiers albeit with fixed sparse~\cite{Babbar17,Jain16,Yen17,Tagami17} or dense features~\cite{Jain19}. These methods attempt to learn a tree~\cite{Prabhu14, Jain16, Prabhu18}, embedding~\cite{Tagami17, Bhatia15} or a brute-force classifier~\cite{Babbar17, Babbar19}. Furthermore, Slice~\cite{Jain19}, Parabel~\cite{Prabhu18b}, and PPD-Sparse~\cite{Yen17} discuss negative sampling in the context of fixed pre-trained features. 

\begin{table}[h!]
    \centering
	\resizebox{\linewidth}{!}
	{
		\begin{tabular}{l|ccccccc}
			\toprule
			\textbf{Dataset} & 
			\makecell{\textbf{Number of}\\\textbf{Free Vectors}} & 
                \makecell{\textbf{Batch}\\\textbf{Size}} & 
			\makecell{$\gamma_{min}$} &
            \makecell{$\gamma_{max}$} &
			\makecell{\textbf{epochs}} & 
			\makecell{\textbf{Learning rate}\\$lr$} & 
			\makecell{\textbf{Maximum seq.}\\\textbf{len} $t_{max}$} \\
			\midrule
			LF-AmazonTitles-131K & 65,536 & 3200 & 0.1 & 0.3 & 300 & 0.0003 & 32 \\
            LF-Amazon-131K & 65,536 & 512 & 0.1 & 0.3 & 300 & 0.0003 & 128 \\
            LF-AmazonTitles-1.3M & 131,072 & 3200 & 0.1 & 0.3 & 300 & 0.0003 & 32\\ 
            LF-WikiTitles-500K & 65,536 & 3200 & 0.1 & 0.3 & 300 & 0.0002 & 32\\
            LF-Wikipedia-500K & 65,536 & 512 & 0.1 & 0.3 & 50 & 0.0001 & 256\\
            LF-WikiSeeAlso-320K & 65,536 & 1024 & 0.1 & 0.3 & 200 & 0.0001 & 128\\

            \bottomrule
    	\end{tabular}
     }
	\caption{\alg's hyper-parameter values on benchmark datasets for reproducibility. Most hyper-parameters were set to their default values across all datasets. \alg samples a single positive per data point. Whereas, \algpp makes use of multiple positives and double the batch size of \alg when possible.}	\label{tab:hyper}
\end{table}

\section{Hyper-parameters} \label{app:hyper_parameters}
\alg's main hyper-parameters are included in Table~\ref{tab:hyper}.
Additionally, we use AdamW optimizer and a weight decay of 0.01 that is removed from Bias, LayerNorm and the bank of free vectors $\mathcal{B}$. We use $\lambda=0.1$ as regularization weight across all the experiments. Moreover, for the Prototype Label Network we use a Transformer Encoder layer (1 attention head, 1024 internal dimensionality, and 0.1 dropout). We tried increased dimensionality and/or number of attention heads observing marginal gains, thus decided to keep it simple.
Additionally, we decided to perform centroid estimation using momentum updates ($\alpha=0.95$ for smooth updates in all our experiments). It is worth noting that we explored other strategies for centroid estimation by: 1) using multiple queries for each label, which led to minor gains (0.1-0.2\%) at the cost of increased computation due to having to encode more query textual embeddings batch-wise; 2) using the exact label centroids in the inference phase, which led to comparable performance.
In general, we set PRIME configurations to fit a single a single 80 GB A100 and \alg can be trained in around 60 hours on LF-AmazonTitles-1.3M dataset. PRIME-lite, instead, can be trained in around 42 hours in the same dataset.
Finally, we use the NGAME sampler to select a subset of negatives and adopt the default values for its cluster size.

\section{Inference} \label{app:inference}
\alg computes and saves the final label prototypes, \ie $\{\vz_l\}_{r=1}^{L}$, once the training is finished. Given a new test query $q$, we follow the steps below: 
\begin{enumerate}
    \item Encode the query $q$ as $\vh_q$.
    \item Compute the scores $\{\vh_q^{\top} \vz_l\}_{r=1}^{L}$. Note that $\vh_q$ and $\{\vz_l\}_{r=1}^{L}$ are L2-normalized.
    \item Sort the label indices on the basis of scores and return top-$k$ most relevant labels.
\end{enumerate}

In practice, we use IndexFlatIP provided by the Faiss~\cite{Johnson19} library. \alg can make predictions within a few milli-seconds on the LF-AmazonTitles-1.3M dataset on a single GPU. It should be noted that the inference complexity of \alg is similar to other XMC algorithms that make predictions on the basis of text embeddings~\cite{Gupta24} or classifiers~\cite{Jain23}. Finally, an Approximate Nearest Neighbor~(ANNS) index~\cite{MalkovY16} can be readily integrated for datasets where label prototypes do not fit on a single GPU.

\section{Sampling} \label{app:sampling}
PRIME samples its positives from a distribution where the probability of sampling a positive is proportional to its inverse propensity score~\cite{Jain16}. In particular, the probabilities for the i-th query $q_i$ are defined as follows:
\begin{equation}
p_{il} = \frac{\gamma_{l}}{\sum_{j \in \cP_i}{\gamma_{j}}}. \nonumber
\end{equation}
Here $\gamma_l$ is the inverse propensity score for the label $l$ (please refer to \cite{Jain16} for more details) and $\cP_i$ is the set of positives for the i-th query data point. Such a positive sampling strategy leads to gains across all evaluation metrics~(please see Table~\ref{tab:ablation_positive_sampling} for results on LF-AmazonTitles-1.3M dataset). It should be noted that the gains are more significant in the propensity score metrics~(PSP@$k$), i.e. more benefits are found on the tail labels. 

\begin{table}
\centering
\resizebox{0.70\linewidth}{!}{
  \begin{tabular}{@{}l|cccccccc@{}}
  \toprule
   \textbf{Positive} & \multirow{2}{*}{\textbf{P@1}} & \multirow{2}{*}{\textbf{P@3}} & \multirow{2}{*}{\textbf{P@5}} & \multirow{2}{*}{\textbf{PSP@1}} & \multirow{2}{*}{\textbf{PSP@3}} & \multirow{2}{*}{\textbf{PSP@5}} & \multirow{2}{*}{\textbf{R@100}} \\
   \textbf{Sampling} & & & & & & \\ \midrule
    Uniform & 57.48	& 49.85	& 44.54	& 29.65	& 34.30	& 36.86 & 62.50 \\ 
    Proposed & 58.10	& 50.39	& 45.05	& 31.68	& 36.12	& 38.55 & 63.45
 \\ 
    \bottomrule
\end{tabular}
}\caption{Impact of positive sampling on the PRIME-lite algorithm on the LF-AmazonTitles-1.3M dataset. The proposed sampling results in prominent gains in the propensity score metrics (PSP@$k$).}
    \label{tab:ablation_positive_sampling}
\end{table}

\end{appendices}

\end{document}